\documentclass[10pt,twocolumn,letterpaper]{article}

\usepackage[pagenumbers]{iccv}      
\usepackage{comment}
\usepackage{amsmath,amsfonts,bm}

\newcommand{\red}[1]{{\color{red}#1}}

\newcommand{\green}[1]{{\color{ForestGreen}#1}}

\usepackage{efbox,graphicx}
\usepackage{caption}
\usepackage{subcaption}
\usepackage{booktabs}
\usepackage{makecell}
\usepackage{pifont}

\usepackage{booktabs, multirow} 
\usepackage{soul}
\usepackage[dvipsnames]{xcolor,colortbl} 
\usepackage{changepage,threeparttable} 
\usepackage{adjustbox}

\def\tick{\ding{51}}
\def\greentick{\green{\ding{51}}}
\def\cross{\red{\ding{55}}}
\definecolor{Gray}{gray}{0.9}
\definecolor{lightblue}{RGB}{224, 242, 241}

\newcommand{\method}{COOkeD}

\usepackage{pythonhighlight}

\efboxsetup{linecolor=black,linewidth=1pt}

\usepackage{stfloats}

\definecolor{iccvblue}{rgb}{0.21,0.49,0.74}
\usepackage[pagebackref,breaklinks,colorlinks,allcolors=iccvblue]{hyperref}

\def\paperID{9} 
\def\confName{ICCV Workshop}
\def\confYear{2025}

\title{COOkeD: Ensemble-based OOD detection in the era of zero-shot CLIP}

\author{Galadrielle Humblot-Renaux$^1$\qquad Gianni Franchi$^2$\qquad Sergio Escalera$^{1,3}$\qquad Thomas B. Moeslund$^1$\\%
$^1$Aalborg University, Denmark\qquad $^2$
ENSTA, Institut Polytechnique de Paris, France\\%
$^3$University of Barcelona and Computer Vision Center, Spain\\%
{\tt\small gegeh@create.aau.dk}\qquad {\tt\small gianni.franchi@ensta.fr}\qquad {\tt\small sescalera@ub.edu}\qquad {\tt\small tbm@create.aau.dk}}

\begin{document}
\maketitle

\begin{abstract}
Out-of-distribution (OOD) detection is an important building block in trustworthy image recognition systems as unknown classes may arise at test-time.  OOD detection methods typically revolve around a single classifier, leading to a split in the research field between the classical supervised setting (e.g. ResNet18 classifier trained on CIFAR100) vs. the zero-shot setting (class names fed as prompts to CLIP). In both cases, an overarching challenge is that the OOD detection performance is implicitly constrained by the classifier's capabilities on in-distribution (ID) data. In this work, we show that given a little open-mindedness from both ends, remarkable OOD detection can be achieved by instead creating a heterogeneous ensemble - COOkeD combines the predictions of a closed-world classifier trained end-to-end on a specific dataset, a zero-shot CLIP classifier, and a linear probe classifier trained on CLIP image features. While bulky at first sight, this approach is modular, post-hoc and leverages the availability of pre-trained VLMs, thus introduces little overhead compared to training a single standard classifier. We evaluate COOkeD on popular CIFAR100 and ImageNet benchmarks, but also consider more challenging, realistic settings ranging from training-time label noise, to test-time covariate shift, to zero-shot shift which has been previously overlooked. Despite its simplicity, COOkeD achieves state-of-the-art performance and greater robustness compared to both classical and CLIP-based OOD detection methods. Code is available at \url{https://github.com/glhr/COOkeD}
\end{abstract}

\section{Introduction}

Classifiers deployed ``in the wild'' may encounter images which do not belong to any classes seen during training and thus which cannot be classified correctly. Many practical applications require these images to be flagged~\cite{ood-medical_2025,ood-insects_2024,ood-robotics_2025} - this is precisely the motivation for out-of-distribution (OOD) detection~\cite{generalized_ood_survey_2024,ood-advances-approaches-2024}. At the system-level, the aim is to classify in-distribution (ID) images accurately while being able to distinguish them from OOD images. 

Traditionally, OOD detection has revolved around standard closed-world classifiers, where a model is trained on a specific ID dataset, and an OOD score is derived using either confidence-based or feature-based methods~\cite{msp-2017,knn-2022,rmds-2021,weiperkld-2024,scale-2024}. At the same time, pre-trained vision-language models (VLMs) are a compelling alternative to standard classifiers, as they enable zero-shot classification out-of-the-box, without needing to train on the ID dataset~\cite{clip-2021}. This has motivated a popular line of research using VLMs for zero-shot OOD detection - in particular, most works in this direction revolve around CLIP~\cite{generalized-ood-clip_2025,recent-advances-clip-ood_2025,ood-limits-2021,zeroshot-ood-clip-2022,mcm_clip_2022,neglabel-2024,clipn-2023,gl_mcm_2025}

\begin{table}[t]
    \centering
    \resizebox{\linewidth}{!}{%
    \begin{tabular}{c|c|c|c|c|c}
        & \makecell{classification\\accuracy}  & \makecell{test-time\\covariate shift} & \makecell{near-\\OOD} & \makecell{zero-shot\\shift} & \makecell{training-time\\label noise} \\ \hline
        \cite{gl_mcm_2025} & \cross &  \cross & \cross & \cross & \multirow{3}{*}{N/A} \\ \
        \cite{zeroshot-ood-clip-2022,mcm_clip_2022,clipn-2023,tag_2024} & \cross  & \cross  & limited & \cross \\ 
        \cite{neglabel-2024} & \cross  & \greentick & limited & \cross \\ \hline
        ours & \greentick & \greentick & \greentick & \greentick & \greentick \\
    \end{tabular}%
    }
    \caption{Zero-shot OOD detection methods are typically evaluated within CLIP's comfort zone, and the classification accuracy is overlooked entirely. Instead, we consider robustness along several axes, both in terms of OOD detection and accuracy.}
    \label{tab:eval_intro}
\end{table}

\begin{figure}[t]
    \centering
    \includegraphics[width=\linewidth]{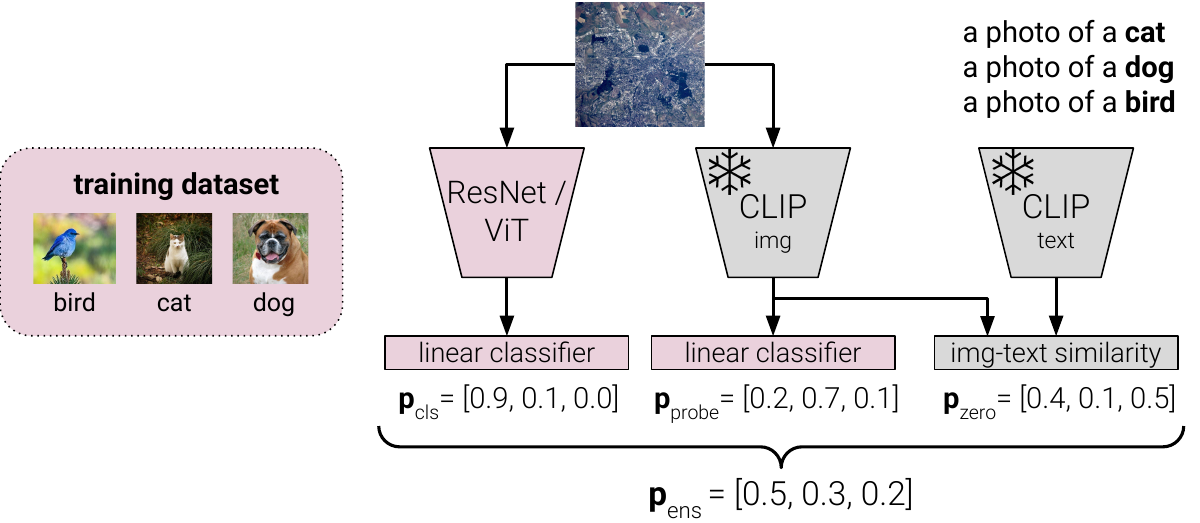}
    \caption{COOkeD in a nutshell, illustrated with a toy example. COOkeD ensembles three models, but only requires a single model to be trained end-to-end thanks to the availability of CLIP.}
    \label{fig:cooked-method}
\end{figure}

The starting point for this work is that both paradigms - closed-world OOD detection and zero-shot CLIP-based approaches - \textbf{have different strengths and limitations}. 

On one hand, CLIP benefits from general semantic knowledge which is independent of the specific ID dataset; CLIP is blind to training data size or quality. However, as summarized in Table~\ref{tab:eval_intro}, the evaluation of CLIP-based OOD detection has thus far been restricted to domains well-aligned with CLIP's ``comfort zone'' where it shows strong zero-shot classification performance (natural object-centric images e.g., ImageNet and related categories~\cite{imagenet-2009})~\cite{neglabel-2024,mcm_clip_2022,gl_mcm_2025}.  Furthermore, while CLIP can excel in the far OOD setting, few works consider the challenge of near OOD shifts. In contrast, \textbf{we consider a more general setting}, where fine-grained semantic shift may occur and \textit{the ID dataset itself may be far from CLIP’s original training distribution} (we call this \textbf{\textit{zero-shot shift}}). In this setting, \textbf{zero-shot CLIP reaches its limits as a standalone solution - thus, training or fine-tuning on the target ID dataset often remains necessary in practice}. 

On the other hand, classical OOD detection methods based on a standard classifier trained on an ID dataset can be used across many domains. However, it has been shown that their performance is tightly linked to the discriminative capabilities of the classifier~\cite{good-closed-set-is-all-you-need-2022,benchmarking-ood-difficulty-2023} and degrades when the training data contains unreliable labels~\cite{ours_noisy-elephant-2024}. As closed-set classifiers struggle with generalization beyond the ID dataset, it is also difficult for OOD detectors in this paradigm to distinguish \textit{covariate-shifted} images of known classes (ID) from \textit{semantically-shifted} images of unseen classes (OOD)~\cite{openood_1.5_2023}.

 In between standard classifiers and zero-shot CLIP, we consider a third classification paradigm: CLIP linear probing, which is a widely adopted baseline for fine-tuning CLIP while keeping its image encoder frozen~\cite{clip-2021}. In ideal settings,  probe CLIP can be a strong contender for OOD detection. We find that, like a standard classifier trained end-to-end, this approach is sensitive to covariate shift, but is less affected by training-time label noise.

Given these tradeoffs, we introduce a new baseline for OOD detection that leverages the strengths of all three approaches. Instead of treating a closed-world classifier, zero-shot CLIP and probe CLIP as mutually exclusive approaches, we propose \textbf{COOkeD} (\textbf{\ul{C}}LIP for \textbf{\ul{OOD}} detection with some extra \textbf{\ul{k}}nowl\textbf{\ul{e}}dge), a framework that integrates zero-shot and task-specific classifiers into an ensemble. This approach enables the use of CLIP as an already-trained model while also incorporating task-specific classifiers. We show that they successfully complement each other for classification and OOD detection, resulting in robust performance across diverse corruptions, shifts and ID domains. Our contributions are as follows:

\begin{itemize}
    \item We explore the limits of relying solely on a standard classifier, zero-shot CLIP or probe CLIP for OOD detection, looking at robustness along several axes which are outlined in Table~\ref{tab:eval_intro}. 
    \item We show the potential of unifying these three classification approaches into a heterogeneous ensemble, introducing a new framework called COOkeD.
    \item We perform ablations validating the contribution of each ensemble member, how to aggregate their predictions and how to derive an OOD score. 
    \item We validate COOkeD in challenging evaluation scenarios, and demonstrate its robustness compared to state-of-the-art methods (both classifier-based and CLIP-based).
\end{itemize}

\section{Background and motivation}

\subsection{Distribution shift and OOD detection}

Given a training dataset $D_{train}$ with a fixed set of $C$ classes,  $D_{train}$ follows a joint distribution $P(X,Y)$, where $X$ represents input variables and $Y$ denotes labels. Given a new image $\textbf{x}$, the goal is to accurately predict its true class. However, during inference in real-world settings, two main types of test-time distribution shifts can be expected:
\begin{itemize}
    \item \textbf{Semantic shift:} The label distribution changes, i.e., $P(Y) \neq P'(Y)$, to the extent that new classes appear in the test data that were absent in $D_{train}$.
    \item \textbf{Covariate shift:} The input distribution shifts, i.e., $P(X) \neq P'(X)$, meaning that known classes from $D_{train}$ appear, but their visual characteristics differ due to variations in lighting, style, quality, or other factors.
\end{itemize}
Alongside the class prediction, the aim of an OOD detector is to compute a score $S(\textbf{x})$ which is a reliable indicator of whether an input is \textit{In-Distribution} (ID) or \textit{Out-of-Distribution} (OOD).
Inputs are considered ID in the absence of semantic shift. In other words, the aim of OOD detection is specifically to identify semantic shift, while covariate-shifted inputs are treated as ID, since the classifier should be able to classify them~\cite{reflect-eval-2023}. 

To evaluate OOD detectors, the scoring function $S(\textbf{x})$ is applied on a test set $D_{test}$ containing only ID images and to a separate $D_{OOD}$ set, containing only OOD images. For effective OOD detection, the OOD score should be consistently lower (or higher) on $D_{test}$ than on $D_{OOD}$, such that OOD samples can be identified via thresholding.

In the choice of $D_{OOD}$, we often differentiate between Near-OOD and Far-OOD scenarios~\cite{openood_1.5_2023}. Near-OOD datasets exhibit only a semantic shift relative to the ID dataset, making them more challenging to detect visually. In contrast, Far-OOD datasets also include a significant covariate shift, which makes them relatively more distinguishable.

\subsection{Classification strategies}

Given the widespread availability of pre-trained CLIP models, we consider three classification approaches around which to design an OOD detector:
\begin{enumerate}
    \item \textbf{Standard Classification:} Train or fine-tune a classifier (e.g., ResNet50) end-to-end on $D_{train}$.
    \item \textbf{Probe CLIP:} Encode $D_{train}$ images using CLIP's (frozen) image encoder and train a linear classifier on the image embeddings.
    \item \textbf{Zero-shot CLIP:} Encode the $C$ ID class names using CLIP's text encoder and classify images based on cosine similarity with text embeddings~\cite{clip-2021}, without requiring any training.
\end{enumerate}
For each of these classification approaches, post-hoc OOD detection has been successfully applied. Despite its simplicity, applying Softmax over the class logits and selecting the Maximum Softmax Probability (MSP) remains a widely used baseline in all three cases~\cite{msp-2017,mcm_clip_2022}.

\subsection{What's the problem?}

When considering a standard classifier, probe CLIP and zero-shot CLIP as three mutually exclusive approaches, we find that \textbf{each approach can excel or struggle}, depending on the combination of $D_{train}$, $D_{test}$ and $D_{OOD}$.

\begin{figure*}[t]
    \centering
    \begin{subfigure}{0.54\textwidth}
    \includegraphics[height=2cm,trim={0 0 0 0.5cm},clip]{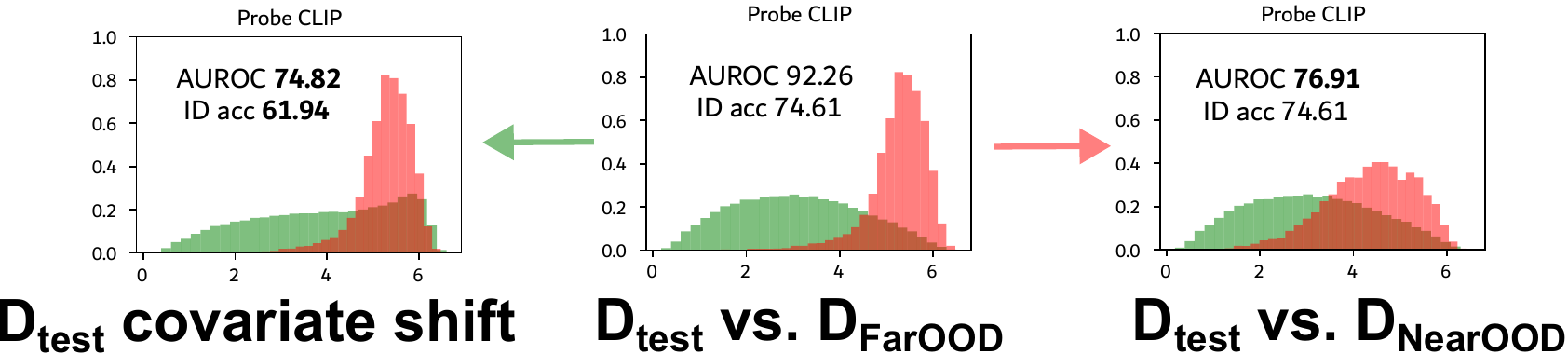}
    \caption{\textbf{Probe CLIP}, with ImageNet1K as $D_{test}$, iNaturalist~\cite{inaturalist-2018} as $D_{FarOOD}$ and NINCO~\cite{ninco-2023} as $D_{NearOOD}$.}
    \label{fig:histos-section2-a}
    \end{subfigure}
    \hfill
    \begin{subfigure}{0.42\textwidth}
    \centering
        \includegraphics[height=2cm,trim={0 0 0 0.5cm},clip]{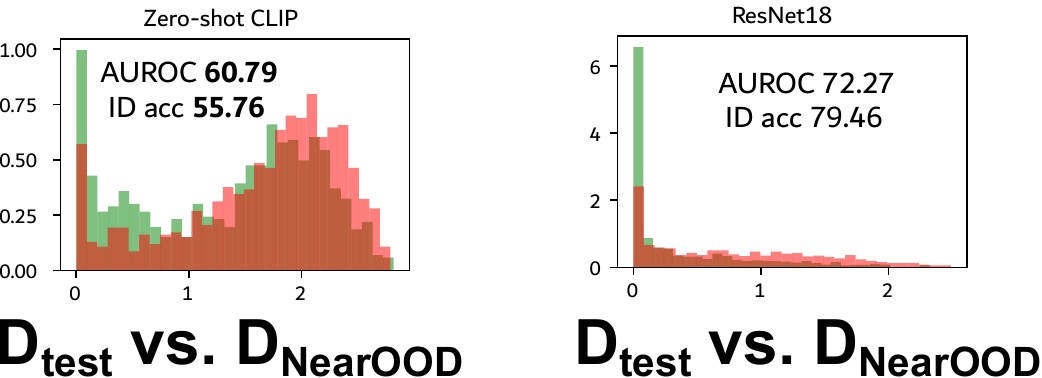}
        \caption{\textbf{Zero-shot CLIP} (left) vs. a fine-tuned \textbf{ResNet-18} (right) with DTD\textsubscript{ID} as $D_{test}$ and DTD\textsubscript{OOD} as $D_{OOD}$~\cite{dtd-2014,ooddb}. }
        \label{fig:histos-section2-b}
    \end{subfigure}
    \caption{Comparison of probe CLIP zero-shot CLIP across different ID/OOD pairs. The plots are histograms of the entropy of the classifier's prediction (OOD score) on the x-axis, and the histogram density on the y-axis. }
    \label{fig:histos-section2}
\end{figure*}

\begin{itemize}
    \item Zero-shot CLIP's performance is agnostic/blind to the labelled examples in $D_{train}$, but is dependent on the set of \textit{classes }present in $D_{train}$ and the images in $D_{test}$, as they may not align with the image-text pairs seen during CLIP's training. For example, while zero-shot CLIP can classify ImageNet-like images with high accuracy, its performance drops significantly on texture or satellite images~\cite{coop-2022}. In the context of OOD detection, we call this problem setting \textbf{\textit{zero-shot shift}}.
    \item A standard classifier or probe CLIP are both sensitive to the size and \textit{contents} of $D_{train}$, which is used to optimize their parameters during training. In particular, label noise in the training data has shown to degrade OOD detection performance~\cite{ours_noisy-elephant-2024,challenges-noisy-nonstandard-classification-2019}. The extent to which $D_{test}$ deviates from $D_{train}$ in terms of covariate shift also hinders OOD detection performance.
    \item The difficulty of a given $D_{OOD}$ is model-dependent. We generally find zero-shot CLIP and probe CLIP to be stronger at far OOD, while a standard classifier has the potential to be a strong near-OOD detector.
\end{itemize}
Figure~\ref{fig:histos-section2} shows examples of challenging scenarios for probe CLIP and zero-shot CLIP. In particular, Figure~\ref{fig:histos-section2} shows that zero-shot CLIP is very poor at near OOD detection when $D_{test}$ is subject to zero-shot shift (unfamiliar domain), and is significantly outperformed by a ResNet-18 trained on texture images in this setting. Our motivation is therefore to \textbf{exploit their complementarity} and their potential as an ensemble of three models. This results in consistent performance across challenging settings (cf. Table~\ref{tab:approach-accuracy-comparison} and Table~\ref{tab:whyclip} for more results).

\section{COOkeD ensemble: our approach}\label{sec:approach}

Figure~\ref{fig:cooked-method} gives a high-level overview of how COOkeD aggregates the predictions of a standard classifier, probe CLIP and zero-shot CLIP. We detail the training procedure and test-time ensembling below.

\subsection{Training}
Given a training dataset $\mathcal{D}_{train} = \{(\mathbf{x}_n, y_n)\}_{n=1}^N$ of images $\mathbf{x}_n \in \mathbb{R}^m$ and their corresponding labels $y_n \in \{1, 2, \ldots, C\}$, we consider a standard deep classifier $f_{cls}: \mathbb{R}^m \to \mathbb{R}^C$ trained end-to-end on $\mathcal{D}_{train}$ to minimize the cross-entropy loss.

In parallel, we consider a CLIP model with an image encoder $E_I:\mathbb{R}^m \to \mathbb{R}^d$ and text encoder $E_T : \mathcal{T} \to \mathbb{R}^d$ which where both trained to align image-text pairs in a $d$-dimensional shared embedding space.

A linear classifier $f_{probe}: \mathbb{R}^d \to \mathbb{R}^C$ defined by weights $\mathbf{W} \in \mathbb{R}^{C \times d}$ and biases $\mathbf{b} \in \mathbb{R}^{C}$ is trained to minimize the cross-entropy loss on the dataset $\{(E_I(\mathbf{x}_n), y_n)\}_{n=1}^N$ while $E_I$ remains frozen.

We place no constraints on the $f_{cls}$ classifier architecture or the CLIP model size/variant. We consider both a standard classifier trained from scratch or a standard classifier initialized from pre-trained ImageNet1K weights.

\subsection{Test-time}

At test-time, classifier predictions are combined at the level of class probabilities, similarly to Deep Ensembles~\cite{deepensembles-2017}. Pseudocode is provided in the Supplementary~\ref{supp:approach}.

\noindent  

\subsubsection{Class logits}

Given an image to classify $\mathbf{x}$, class logits $\textbf{l}$ can directly be obtained from the standard classifier  $\textbf{l}_{cls} = f_{cls}(\mathbf{x})$ and and linear probe $\textbf{l}_{probe} = \mathbf{W} E_I(\mathbf{x}) + \mathbf{b}$.

For classification using zero-shot CLIP, a set of text prompts $ \{p_c\}_{c=1}^C $ is first constructed, where each prompt $p_c$ takes the format ``a photo of a [cls]'' (e.g. ``a photo of a cat''). These prompts are then tokenized and embedded using the text encoder $E_T$, resulting in text embeddings $ \{E_T(p_c)\}_{c=1}^C$. This step only needs to be performed once.

To obtain zero-shot class logits $ \textbf{l}_{zero}$ for $\mathbf{x}$, the cosine similarity between $E_I(\mathbf{x})$ and each class' text embedding $E_T(p_c)$ is computed and scaled by CLIP's temperature $\tau$ (a learned parameter): $ \textbf{l}_{zero} = \left[ \tau \cos(E_I(\mathbf{x}), E_T(p_1)), ..., \tau \cos(E_I(\mathbf{x}), E_T(p_C)) \right] $.

\subsubsection{Ensembling}

Class logits $\textbf{l}_{cls}$, $\textbf{l}_{probe}$ and $\textbf{l}_{zero}$ are first softmax-normalized to obtain class probability vectors $\textbf{p}_{cls}$, $\textbf{p}_{probe}$ and $\textbf{p}_{zero}$ of the form $\textbf{p} =  [p(y=1|\mathbf{x}),...,p(y=C|\mathbf{x})] $. Class probabilities are then averaged across the ensemble members, such that the probability of class $c$ is:
\[p_{ens}(y=c|\mathbf{x}) = \frac{1}{|M|} \sum_{m \in M} p_m (y=c|\mathbf{x})\]
with $M = \{cls,probe,zero\}$.

We can consider different classifier combinations (e.g. $M = \{cls,probe\}$ or $M = \{probe,zero\}$ or $M = \{cls,zero\}$) with different cost-performance tradeoffs. 

\subsubsection{Classification and OOD score}
The predicted class $\hat{y}$ is taken as:
\[\hat{y} = \mathop{\mathrm{arg\,max}}(p_{ens}(y=1|\textbf{x}),...,p_{ens}(y=C|\textbf{x}))\]
As OOD scoring function, we consider the MSP and the categorical entropy, two standard uncertainty measures~\cite{deepensembles-2017}:
\begin{align*}
\mathrm{MSP}(\textbf{p}_{ens}) &= \max(p_{ens}(y=1|\textbf{x}),...,p_{ens}(y=C|\textbf{x})) \\
H(\textbf{p}_{ens}) &= - \sum_{c=1}^C p(y=c|\textbf{x}) \log(p(y=c|\textbf{x}))
\end{align*}

\section{Related work}

\subsection{Classifier-based OOD detection}

We refer to OpenOOD~\cite{openood_1.0_2023,openood_1.5_2023}, a popular OOD detection benchmark, for a detailed categorization. Post-hoc methods assume an off-the-shelf already trained closed-world classifier, while training-based methods modify the classification objective to improve OOD detection.  Our approach falls under the posthoc category and adopts a logit-based OOD score. It can be applied to any classifier - however our evaluation specifically considers a standard classifier trained with a Cross-Entropy objective. Similarly to MSP~\cite{msp-2017} and MaxLogit~\cite{mls-2022}, COOkeD does not rely on outlier examples or training/validation-data statistics, and does not introduce any hyper-parameters. 

\subsection{Zero-shot OOD detection with CLIP}

Recent work has shown the potential of CLIP for zero-shot OOD detection, bypassing the need for training on a specific image classification dataset~\cite{recent-advances-clip-ood_2025,generalized-ood-clip_2025}. 

In this direction, some methods consider the ID classes as the only source of knowledge. As a baseline,~\cite{ood-limits-2021} applies softmax directly on the joint image-text embedding and takes the sum of class-wise probabilities as OOD score. MCM~\cite{mcm_clip_2022} instead applies an MSP score, and shows that the softmax temperature plays a key role in ID vs. OOD separability. GL-MCM~\cite{gl_mcm_2025} incorporates local alignment features to account for the potential presence of OOD objects in ID images. TAG~\cite{tag_2024} implements other post-hoc scoring methods (e.g. REACT, KNN) on CLIP's joint image-text embedding and shows that it can be beneficial to average OOD scores across multiple augmented/shuffled prompts.

Another branch of methods exploits or infers OOD labels. The idea is that after normalizing the joint image-text embedding, probabilities will be lower on the ID classes if OOD classes are added to the set of candidate prompts. ~\cite{ood-limits-2021} showed that strong OOD detection performance can be achieved on CIFAR benchmarks by directly using the name of OOD classes as text prompts for CLIP - however, assuming prior knowledge of all OOD classes can be considered an unfair advantage. ZOC~\cite{zeroshot-ood-clip-2022} trains a separate text encoder on the COCO dataset to produce candidate prompts from CLIP image embeddings. This can then be used to extract candidate words describing a (potentially OOD) input image. NegLabel~\cite{neglabel-2024} proposes a training-free approach for creating negative prompts, selecting words from the WordNet database most dissimilar to the ID class names. 

Our approach most closely aligns with MCM in terms of how the OOD score is computed. It uses CLIP ``as-is'' without training on additional text encoder, and does not rely on negative prompts or candidate OOD labels. However, it breaks free from the ``zero-shot'' paradigm by leveraging classifiers trained on the ID dataset alongside zero-shot CLIP, to boost the classification and OOD detection performance.

\subsection{Model ensembles}

Ensembling consists of aggregating multiple predictions to improve overall predictive performance, robustness and/or uncertainty estimation.

\paragraph{What to combine:} While there are other ways of achieving predictive diversity (e.g. test-time input perturbations), the focus of our work is model ensembling. Classically, ensembles are formed by several models trained on the same dataset and with the same training objective, but with different random initializations~\cite{deepensembles-2017} or hyperparameters~\cite{hyperparamensembles-2020}. More recently,~\cite{mc-ensemble-2024} shows that ensembling models trained under different objectives improves representation diversity and is beneficial for OOD detection. Going a step further, our approach creates a heterogeneous ensemble of models with different (pre-)training objectives, training datasets and architectures.  To the best of our knowledge, the idea of ensembling a standard classifier and a zero-shot and/or probe CLIP has not been explored in prior work.

\paragraph{How to combine:} 
~\cite{mc-ensemble-2024} combines at the feature level (before the prediction head). However, this requires that all models in the ensemble have the same feature dimensionality. There have also been works combining/averaging model weights~\cite{model_ratatouille-2023}. This requires compatible architectures within the ensemble. We combine at the level of class probabilities, similarly to Deep Ensembles~\cite{deepensembles-2017}. This allows for heterogeneous model sizes and architectures within the ensemble (for example, a large CLIP model can be used alongside a lightweight ResNet18 classifier).

\section{Experimental set-up}~\label{sec:xp-setup}

Our experimental set-up is design to align with existing OOD evaluation benchmarks while also incorporating additional challenges. Details for each sub-section can be found in the Supplementary~\ref{supp:xp-setup}.

\subsection{Datasets}\label{sec:setup-datasets}

\textbf{Standard datasets} For direct comparison with existing methods on the OpenOOD leaderboard, we consider CIFAR100~\cite{cifar-dataset-2009}, ImageNet200~\cite{openood_1.5_2023} and ImageNet1K~\cite{imagenet-2009,ILSVRC15} as ID datasets. We follow the OpenOOD benchmark~\cite{openood_1.0_2023,openood_1.5_2023} and data splits, which defines a set of near and far OOD datasets as follows:
\begin{itemize}
    \item CIFAR100 - near: CIFAR10~\cite{cifar-dataset-2009}, TIN~\cite{tinyimagenet-2015} - far: MNIST~\cite{mnist-dataset-2012}, SVHN~\cite{svhn-dataset-2011}, Textures~\cite{textures-2014}, Places365~\cite{places365-dataset-2017}.
    \item ImageNet200 \& ImageNet1K - near: NINCO~\cite{ninco-2023}, SSB-hard~\cite{good-closed-set-is-all-you-need-2022} - far: iNaturalist~\cite{inaturalist-2018}, Textures~\cite{textures-2014}, OpenImage-O~\cite{vim-2022}. 
\end{itemize}

\noindent Following the \textit{full-spectrum} OpenOOD benchmark, we also consider the \textit{covariate-shifted} setting, where at test-time, the ImageNet200 \& ImageNet1K ID images are combined with ImageNet-like images affected by corruptions (ImageNet-C~\cite{imagenet_c-2019}), different styles (ImageNet-R~\cite{imagenet_r-2021}) or increased difficulty (ImageNet-V2~\cite{imagenet_v2}). 

In the \textit{label noise} setting studied by~\cite{ours_noisy-elephant-2024}, CIFAR100-N~\cite{cifar-noisy-dataset-2022} (a re-annotation of CIFAR100's training set with $~\approx40\%$ mislabelled images - is used for training). At test-time, the evaluation is identical to CIFAR100. 

\noindent\textbf{Beyond CIFAR and ImageNet} Lastly, to shift away from CLIP's ``comfort zone'' and introduce \textit{zero-shot shift}, we consider two ID datasets from the OODDB Benchmark~\cite{ooddb}: PatternNet~\cite{patternnet_2018} (satellite images categorized into 38 land types/makrks) and Describable Textures~\cite{dtd-2014} (textural images categorized into 47 descriptions). For these two datasets, OODDB defines a set of held-out classes which we treat as near OOD. As far OOD, we use similar ones as for CIFAR100 above:

\begin{itemize}
    \item PatternNet$_{\texttt{ID}}$ - near: PatternNet$_{\texttt{OOD}}$ - far: CIFAR10, TIN, MNIST, SVHN, DTD$_{\texttt{OOD}}$, Places365.
    \item DTD$_{\texttt{ID}}$ - near: DTD OODDB - far: CIFAR10, TIN, MNIST, SVHN, PatternNet$_{\texttt{OOD}}$, Places365.
\end{itemize}

\begin{table*}[b]
    \centering
    \begin{subtable}{0.47\linewidth}
    \resizebox{\linewidth}{!}{%
    \begin{tabular}{ccccc} \toprule
      \makecell{train / test\\ ID datasets} & \makecell{Standard\\classifier} & \makecell{Probe\\CLIP} & \makecell{Zero-shot\\CLIP} & \makecell{COOkeD\\(ours)}\\ \midrule
      CIFAR-100 & \green{83.30} & 79.98 & 66.50 & \textbf{86.16} \\
      \textbf{CIFAR-100-N} / CIFAR-100 & \red{62.22} & \green{71.21} & 66.50 & \textbf{74.60} \\ \midrule

      PatternNet (19 classes) & \green{99.78} & 99.59 & \red{70.12} & \textbf{99.86} \\
      DTD (23 classes) & 79.46 & \green{83.70} & \red{55.76} & \textbf{84.57} \\ \midrule

      ImageNet-1K & 69.23 & \green{74.61} & \red{67.01} & \textbf{77.26} \\
      ImageNet-1K / \textbf{ImageNet-1K-CS} & \red{48.77} & \green{61.94} & 59.22 & \textbf{64.67} \\ \bottomrule
    \end{tabular}%
    }
    \caption{Classification accuracy}
    \end{subtable}\hfill
    \begin{subtable}{0.516\linewidth}
    \resizebox{\linewidth}{!}{%
    \begin{tabular}{ccccc} \toprule
      \makecell{near / far\\ OOD datasets} & \makecell{Standard\\classifier} & \makecell{Probe\\CLIP} & \makecell{Zero-shot\\CLIP} & \makecell{COOkeD\\(ours)}\\ \midrule
      CIFAR-10 / SVHN & \green{83.16} / \red{81.25} & 79.90 / \green{93.30} & \red{73.99} / 82.19 & \textbf{84.32} / \textbf{93.75} \\
      CIFAR-10 / SVHN & \red{71.56} / \red{81.35} & 74.55 / \textbf{\green{92.79}} & 73.99 / 82.19 & \textbf{78.00} / 92.51 \\ \midrule

      PatternNet\textsubscript{OOD} / MNIST & 94.91 / 99.78 & \textbf{\green{96.32}} / \green{99.94} & \red{72.50} / \red{90.56} & 94.30 / \textbf{100.0} \\
      DTD\textsubscript{OOD} / MNIST & 71.53 / 86.66 & \textbf{\green{80.74}} / \green{96.40} & \red{59.63} / \red{85.96} & 77.70 / \textbf{97.33} \\ \midrule

      NINCO / iNaturalist & \green{77.66} / \red{87.06} & 76.92 / \green{89.07} & \red{75.06} / 88.54 & \textbf{82.22} / \textbf{93.01} \\
      NINCO / iNaturalist & 63.91 / 76.25  & \red{60.40} / \red{74.91} & \green{66.17} / \green{81.90} &  \textbf{68.12} / \textbf{83.79} \\ \bottomrule
    \end{tabular}%
    }
    \caption{OOD detection AUROC, using the MSP as OOD score}
    \end{subtable}

    \caption{Performance (\%) across different datasets, using a fine-tuned ResNet18 as standard classifier and a ViT-B-16 CLIP. With CIFAR-100-N, the classifier and probe are trained on \textbf{noisy labels}. With ImageNet-1K-CS, the test set includes images with \textbf{covariate shift}. We refer to Section~\ref{sec:xp-setup} for details. The best/worst performance across individual classifiers is in green/red. Best overall performance is in bold.}
    \label{tab:approach-accuracy-comparison}
\end{table*}

\subsection{Models and training}

\textbf{Standard classifiers:} We consider several architectures: ResNet18 trained from scratch, ResNet18 from pre-trained, ResNet50 from pre-trained~\cite{resnet-2016} and ViT-B-16~\cite{vit-2021} from pre-trained. For ImageNet1K, we directly use available models from torchvision~\cite{torchvision2016}. For the rest of the ID datasets, classifiers are trained following the procedure and hyperparameters used in the OpenOOD benchmark for training (100 epochs, SGD with momentum and weight decay, cosine annealing). The initial learning rate is set to 0.1 when training from scratch or 0.01 when fine-tuning from a torchvision checkpoint~\cite{torchvision2016}. In all cases, the final checkpoint (after training for 100 epochs) is used for evaluation.

\textbf{Zero-shot and probe CLIP:}  We consider different CLIP models: ViT-B-32, ViT-B-16, ViT-L-14 from OpenAI~\cite{clip-2021} and ViT-H-14 with LAION weights~\cite{open_clip-2023}. For zero-shot classification, the models are used off-the-shelf with the learned temperature ($\tau = 100$ for all CLIP models). For linear probing, we train a single fully connected layer for 20 epochs with the same hyper-parameters as above (initial learning rate of 0.1).

\subsection{Evaluation metrics}

OOD performance is evaluated in terms of AUROC for each OOD dataset. FPR\@95 results are reported in Appendix~\ref{supp:results}. Performance is aggregated into \textit{near OOD}, \textit{far OOD} and \textit{avg. OOD} (average of the near and far OOD performance). The performance of COOkeD on individual OOD datasets is reported in Appendix~\ref{supp:results}.

\section{\method{} in action}

\subsection{Preheating the oven}

Table~\ref{tab:approach-accuracy-comparison} highlights the limitations of relying on a single classifier flavor for classification and OOD detection (echoing the examples from Figure~\ref{fig:histos-section2}), and shows COOkeD's performance in comparison. By simply ensembling the three classifiers, classifier- or dataset-specific performance drops are greatly reduced.

\subsection{To MSP or not to MSP?} 
\begin{figure}[t]
    \centering
    \includegraphics[width=0.8\linewidth]{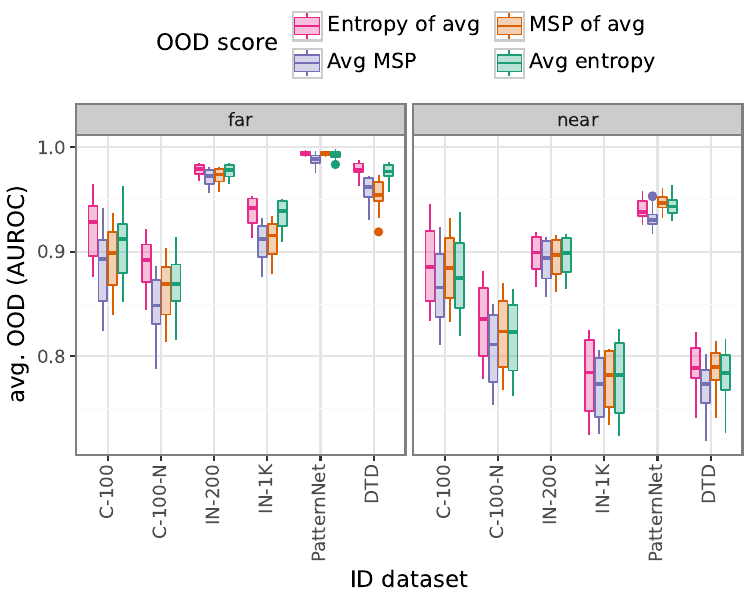}
    \caption{Near vs. far OOD detection with a [cls+probe+zero] ensemble for different OOD scoring functions. Boxplots show the AUROC distribution across 4 classifier and 4 CLIP variants.}
    \label{fig:ood_ablation_oodscore}
\end{figure}

The MSP is widely used in the OOD detection literature and only captures the confidence of the top class, while the entropy describes the full predictive distribution. Orthogonally, the averaging across ensemble predictions can be performed either before or after applying the OOD scoring function. Figure~\ref{fig:ood_ablation_oodscore} shows that an entropy-based score improves OOD detection in the far-OOD setting compared to MSP. It also yields lower FPR95 on all datasets except PatternNet (cf. Appendix Figure~\ref{fig:supp_ood_ensemble_combinations}). Averaging the MSP of individual classifiers in the ensemble consistently gives the worst performance (this is unsurprising, as it does not capture uncertainty arising from disagreement). The rest of the results are reported with the entropy of the averaged class probabilities as OOD score, as described in Section~\ref{sec:approach}.

\begin{figure}[t]
    \centering
     \includegraphics[width=1\linewidth,trim={0 3cm 0 0},clip]{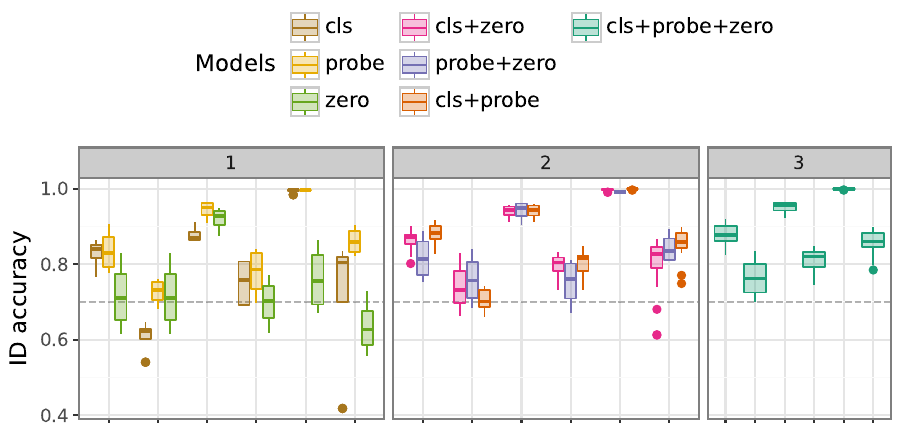}
    \includegraphics[width=1\linewidth,trim={0 0 0 3cm},clip]{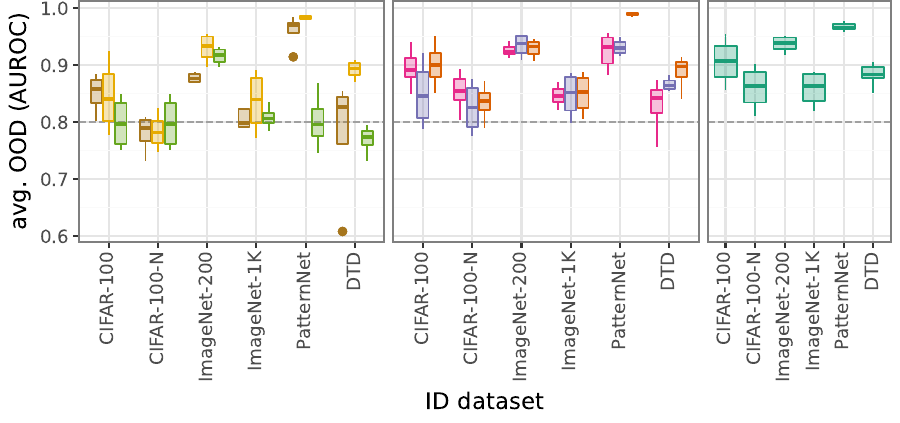}
    \caption{Comparison of different combinations of Zero-shot / Probe / Classifier predictions in terms of classification (top) and OOD detection performance (bottom). Boxplots show the distribution in performance across the 4 CLIP and 4 classifier variants. The plot is split based on the number of classifiers in the ensemble (1,2, or 3).}
    \label{fig:ood_ensemble_combinations}
\end{figure}

\subsection{Does everyone contribute to the ensemble?}
Figure~\ref{fig:ood_ensemble_combinations} compares different ensemble combinations with single-classifier baselines, both in terms of accuracy and OOD detection. We refer to the standard classifier as ``cls'', probe CLIP as ``probe'' and zero-shot CLIP as ``zero''; ``cls+zero'' for example refers to an ensemble of a standard classifier with zero-shot CLIP. Figure~\ref{fig:ood_ensemble_combinations} shows that the main benefit to ensembling is that it brings consistency across datasets. Using cls+probe+zero keeps ID accuracy above 75\% and AUROC above 80\% (dashed lines).

\subsection{Why even ensemble with CLIP?}

Why mix supervised and zero-shot models? What about ensembles of multiple supervised models or multiple CLIPs? Table~\ref{tab:whyclip} compares different ensembling strategies (with 2 or 3 models) on the three most challenging datasets, not only in terms of performance, but also in terms of training cost.

In the absence of training-time label noise (CIFAR100-N) or test-time covariate shift (ImageNet-1K CS), a multi-architecture ensemble of standard classifiers can achieve solid performance. In comparison, \textbf{COOkeD performs more consistently and at a much lower training cost}. Only relying on CLIP (probe or zero-shot), while very efficient in terms of training (or lack thereof), means inheriting their respective weaknesses, and sometimes performing \textbf{worse than a single ResNet18 model} (first model in Table~\ref{tab:whyclip}).

The last section of Table~\ref{tab:whyclip} shows how COOkeD's performance can be pushed further by using a larger CLIP model. Note that this comes at no additional training cost with a cls+zero ensemble, and with only a small training cost with cls+probe+zero, as the input dimensionality of the probe is increased.

\begin{table}[t]
    \centering
    \resizebox{\linewidth}{!}{%
    \begin{tabular}{rcccccc}
    \toprule
    ID dataset $\rightarrow$
    & CIF100-N
    & DTD
    & \multicolumn{2}{c}{ImageNet-1K}
    & \multirow{2}{*}{\makecell{train\\param}}\\

    cov. shift (CS)? $\rightarrow$ & & & & \tick \\
    \midrule
        single R18
        & \red{62.22} / \red{77.90}
        & \red{79.46} / \red{81.23}
        & \red{69.23} / \red{79.24}
        & \red{48.77} / \red{64.50}
        & 11.7M\\
        
        R18$\times$3 
        & \red{66.29} / 81.13
        & \red{80.98} / \red{83.34}
        & - & - & 35.1M\\
       
        CLIP$\times$3
        & 76.81 / 81.49
        & \red{68.37} / \red{79.42}
        & \red{73.97} / 83.26
        & 69.18 / 74.76
        & N/A\\
        \midrule
        \multicolumn{6}{c}{Ensembling with ViT-B-16 (trained on ID) vs. CLIP-B-16  $\downarrow$} \\ \midrule

        R18+ViT
        & \red{65.94} / 83.52
        & 82.61 / 86.27
        & 79.88 / \red{82.93}
        & \red{60.27} / \red{66.97}
        & 98.3M\\
        
        R50+ViT
        & 66.82 / 85.14
        & 83.37 / 87.11
        & 81.00 / 84.35
        & 61.11 / 67.80
        &112.1M\\

        R18+R50+ViT
        & 69.16 / 85.03
        & 83.37 / 85.82
        & 79.97 / 84.00
        & \red{60.02} / 67.04
        &123.8M\\

        \rowcolor{Gray} R18+CLIP
        & 72.08 / 83.81
        & 80.76 / 82.35
        & 75.64 / 83.50
        & 62.67 / 70.14
        & 11.7M\\
        
        \rowcolor{Gray} \textbf{Probe+R18+CLIP}
        & 74.60 / 84.04
        & 84.57 / 87.95
        & 77.26 / 84.08
        & 64.67 / 68.70
        & 12.2M\\

        \rowcolor{Gray} R50+CLIP
        & 73.01 / 85.16
        & 82.07 / 83.52
        & 78.81 / 84.79
        & 64.68 / 70.77
        & 25.6M\\
        
        \rowcolor{Gray} \textbf{Probe+R50+CLIP}
        & 75.11 / 84.94
        & 85.22 / 88.63
        & 79.76 / 84.88
        & 66.17 / 69.04
        & 26.1M  \\

        \rowcolor{lightblue} Probe only
        & 71.21 / \red{76.91}
        & 83.70 / 88.71
        & \red{74.61} / \red{80.72}
        & 61.94 / \red{62.12}
        & 0.5M \\

        \midrule
        \multicolumn{6}{c}{Using CLIP-L-14 instead $\downarrow$} \\ \midrule

        \rowcolor{Gray} R-18+CLIP
        & 76.88 / 87.47
        & 83.04 / 84.48
        & 78.57 / 85.80
        & 70.27 / 74.79
        & 11.7M\\

        \rowcolor{Gray} \textbf{Probe+R18+CLIP}
        & \textbf{79.56} / \textbf{88.13}
        & \textbf{87.83} / \textbf{89.13}
        & \textbf{81.83} / \textbf{87.80}
        & \textbf{73.73} / \textbf{75.18}
        & 12.5M\\

        \rowcolor{Gray} R-50+CLIP
        & \textbf{77.34}/ \textbf{88.41}
        & 84.89 / 85.54
        & 80.79 / 87.09
        & 71.37 / \textbf{75.40}
        & 25.6M\\
        
        \rowcolor{Gray} \textbf{Probe+R50+CLIP}
        & \textbf{79.48} / \textbf{88.78}
        & \textbf{88.15} / \textbf{89.72}
        & \textbf{82.99} / \textbf{88.48}
        & \textbf{74.33} / \textbf{75.44}
        & 26.3M  \\

        \rowcolor{lightblue} Probe only
        & 75.04 / \red{79.46}
        & \textbf{88.15} / \textbf{90.20}
        & \textbf{82.64} / \textbf{87.23}
        & \textbf{73.94} / 72.04
        & 0.8M\\
        
        \bottomrule
        \end{tabular}%
    }
    \caption{ID classification accuracy / OOD detection AUROC (\%) for different ensembling strategies. CLIP$\times$3 is an ensemble of CLIP-B32+CLIP-B16+CLIPL14. The number of training params. is calculated for 1000 classes. \red{Worst 3} and \textbf{best 3} are highlighted.}
    \label{tab:whyclip}
\end{table}

\section{Benchmarking}
\subsection{Comparison to CLIP-based OOD detection}

We compare to ZOC~\cite{zeroshot-ood-clip-2022}, MCM~\cite{mcm_clip_2022}, GL-MCM~\cite{gl_mcm_2025}, TAG~\cite{tag_2024} and NegLabel~\cite{neglabel-2024} (we also considered CLIPN~\cite{clipn-2023} but were unable to reproduce the paper's results). Results are in Table~\ref{tab:comparison-clipbbased} for different CLIP architectures. We find that across all baseline methods, there is a wide gap between near and far OOD performance. Distinguishing DTD vs. near OOD images is the most challenging, with none of the baselines exceeding 70\% AUROC. This shows the importance of considering zero-shot shift when evaluating CLIP-based OOD detection. ImageNet200 and ImageNet1K share the same OOD datasets, yet there is a 10+\% gap in near OOD detection performance between the two, showing the difficulty introduced by increasing the size of the ID label space.    

Interestingly, generating OOD prompts (as done by ZOC and NegLabel) does not bring any performance gains under \textit{zero-shot shift} - in fact, under zero-shot shift, NegLabel often performs \textit{worse} than MCM, especially in near OOD. 

In contrast, by bringing in a classifier trained on the ID dataset, COOkeD narrows the gap between near and far OOD detection and achieves consistent performance even under \textit{zero-shot shift} (PatternNet and DTD). 

\begin{table*}[!htp]\centering
\resizebox{\textwidth}{!}{%

\begin{tabular}{cccrccccccccccccccccccccc}\toprule
& & &ID dataset $\rightarrow$ & &\multicolumn{3}{c}{PatternNet\textsubscript{ID}} & &\multicolumn{3}{c}{DTD\textsubscript{ID}} & &\multicolumn{3}{c}{CIFAR100} & &\multicolumn{3}{c}{ImageNet200} & &\multicolumn{3}{c}{ImageNet1K} \\\cmidrule{4-4}\cmidrule{6-8}\cmidrule{10-12}\cmidrule{14-16}\cmidrule{18-20}\cmidrule{22-24}
\makecell{method \\/ paper} & \makecell{extra\\model} &\makecell{neg.\\prompts} &OOD detector & & \makecell{near\\OOD} &\makecell{far\\OOD} &\makecell{avg\\OOD} & &\makecell{near\\OOD} &\makecell{far\\OOD} &\makecell{avg\\OOD} & &\makecell{near\\OOD} &\makecell{far\\OOD} &\makecell{avg\\OOD} & &\makecell{near\\OOD} &\makecell{far\\OOD} &\makecell{avg\\OOD} & &\makecell{near\\OOD} &\makecell{far\\OOD} &\makecell{avg\\OOD} \\\midrule

\multicolumn{24}{c}{ViT-B-32 $\downarrow$} \\ \midrule

MCM~\cite{mcm_clip_2022} & & &MSP ($\tau=1$) & &66.26 &84.82 &\cellcolor[HTML]{e67c73}75.54 & &63.19 &91.57 &\cellcolor[HTML]{fbd693}77.38 & &68.75 &80.03 &\cellcolor[HTML]{efa581}74.39 & &83.29 &96.37 &\cellcolor[HTML]{f6c08b}89.83 & &68.06 &90.15 &\cellcolor[HTML]{f4ba89}79.11 \\
GL-MCM~\cite{gl_mcm_2025} & & &GL-MCM & &64.62 &86.33 &\cellcolor[HTML]{e67c73}75.48 & &63.43 &90.85 &\cellcolor[HTML]{fbd492}77.14 & &68.73 &79.73 &\cellcolor[HTML]{efa481}74.23 & &83.83 &96.26 &\cellcolor[HTML]{f7c48d}90.05 & &69.41 &90.06 &\cellcolor[HTML]{f6c38c}79.74 \\
TAG~\cite{tag_2024} & & &MSP +TAG & &67.98 &84.20 &\cellcolor[HTML]{e78174}76.09 & &60.29 &88.93 &\cellcolor[HTML]{f4b989}74.61 & &66.89 &80.29 &\cellcolor[HTML]{ee9f7f}73.59 & &83.08 &94.71 &\cellcolor[HTML]{f2b085}88.89 & &67.59 &85.05 &\cellcolor[HTML]{ec977c}76.32 \\
TAG~\cite{tag_2024} & & &MaxLogit+TAG & &70.75 &97.30 &\cellcolor[HTML]{f7c58d}84.02 & &58.26 &91.21 &\cellcolor[HTML]{f4ba89}74.73 & &78.54 &88.26 &\cellcolor[HTML]{fee398}83.40 & &81.54 &94.64 &\cellcolor[HTML]{efa280}88.09 & &65.93 &85.80 &\cellcolor[HTML]{eb917a}75.86 \\
ZOC~\cite{zeroshot-ood-clip-2022} &BERT &\tick &MSP & &68.66 &96.97 &\cellcolor[HTML]{f5bb89}82.81 & &58.40 &90.15 &\cellcolor[HTML]{f3b587}74.28 & &78.56 &86.87 &\cellcolor[HTML]{fddf96}82.72 & &80.81 &90.89 &\cellcolor[HTML]{e67c73}85.85 & &66.86 &81.57 &\cellcolor[HTML]{e67c73}74.22 \\
NegLabel~\cite{neglabel-2024} & &\tick & NegLabel & &57.81 &97.16 &\cellcolor[HTML]{ea8d79}77.48 & &52.09 &89.78 &\cellcolor[HTML]{eb927a}70.94 & &72.75 &80.12 &\cellcolor[HTML]{f3b387}76.43 & &83.50 &97.35 &\cellcolor[HTML]{f8ca8f}90.42 & &70.65 &93.66 &\cellcolor[HTML]{fee297}82.15 \\

\rowcolor{Gray}\textbf{ours} &\textbf{} & &[cls+clip] & &83.14 &95.04 &\cellcolor[HTML]{e7df97}89.09 & &72.76 &96.52 &\cellcolor[HTML]{a5cf91}84.64 & &85.46 &90.43 &\cellcolor[HTML]{aed192}87.95 & &87.05 &96.73 &\cellcolor[HTML]{fee398}91.89 & &75.26 &91.68 &\cellcolor[HTML]{e1de97}83.47 \\

\rowcolor{Gray}\textbf{ours} &\multirow{-2}{*}{R-50} & &[cls+clip+probe] & &92.68 &99.04 &\cellcolor[HTML]{74c38d}95.86 & &78.58 &97.84 &\cellcolor[HTML]{6fc18d}88.21 & &85.01 &88.68 &\cellcolor[HTML]{c3d694}86.84 & &87.20 &97.04 &\cellcolor[HTML]{f7e399}92.12 & &73.63 &91.89 &\cellcolor[HTML]{f5e399}82.76 \\
\midrule

\multicolumn{24}{c}{ViT-B-16 $\downarrow$} \\ \midrule

MCM~\cite{mcm_clip_2022} & & &MSP ($\tau=1$) & &71.85 &86.81 &\cellcolor[HTML]{ed9d7f}79.33 & &61.84 &86.91 &\cellcolor[HTML]{f3b688}74.37 & &73.63 &79.10 &\cellcolor[HTML]{f3b386}76.37 & &84.60 &96.90 &\cellcolor[HTML]{fad091}90.75 & &68.90 &91.33 &\cellcolor[HTML]{f8c78e}80.12 \\
GL-MCM & & &GL-MCM & &73.98 &91.30 &\cellcolor[HTML]{f4b989}82.64 & &61.87 &86.97 &\cellcolor[HTML]{f4b788}74.42 & &75.22 &83.35 &\cellcolor[HTML]{f7c78e}79.28 & &85.77 &97.13 &\cellcolor[HTML]{fcdc95}91.45 & &71.03 &92.01 &\cellcolor[HTML]{fcd995}81.52 \\
TAG~\cite{tag_2024} & & &MSP+TAG & &71.41 &86.15 &\cellcolor[HTML]{ec987d}78.78 & &57.39 &85.28 &\cellcolor[HTML]{ec967c}71.34 & &67.47 &78.39 &\cellcolor[HTML]{ed9b7e}72.93 & &85.98 &95.79 &\cellcolor[HTML]{fad292}90.89 & &69.93 &87.61 &\cellcolor[HTML]{f3b688}78.77 \\
TAG~\cite{tag_2024} & & &MaxLogit+TAG & &71.78 &97.18 &\cellcolor[HTML]{f8c98f}84.48 & &58.36 &79.39 &\cellcolor[HTML]{e67c73}68.87 & &74.62 &83.25 &\cellcolor[HTML]{f7c48d}78.94 & &82.88 &94.77 &\cellcolor[HTML]{f2af85}88.82 & &66.33 &86.43 &\cellcolor[HTML]{ec977d}76.38 \\
NegLabel & &\tick &NegLabel & &65.85 &98.49 &\cellcolor[HTML]{f3b587}82.17 & &55.03 &90.67 &\cellcolor[HTML]{f0a682}72.85 & &70.10 &66.77 &\cellcolor[HTML]{e67c73}68.44 & &86.08 &97.96 &\cellcolor[HTML]{fce599}92.02 & &73.15 &94.65 &\cellcolor[HTML]{d6db96}83.90 \\

\rowcolor{Gray}\textbf{ours} &\textbf{} & &[cls+clip] & &85.25 &95.60 &\cellcolor[HTML]{d1da95}90.42 & &71.93 &95.11 &\cellcolor[HTML]{b6d393}83.52 & &87.23 &90.68 &\cellcolor[HTML]{9bcc91}88.96 & &88.39 &97.22 &\cellcolor[HTML]{d3da96}92.80 & &76.72 &92.86 &\cellcolor[HTML]{bdd594}84.79 \\

\rowcolor{Gray}\textbf{ours} & \multirow{-2}{*}{R-50} & &[cls+clip+probe] & &92.54 &99.28 &\cellcolor[HTML]{73c28d}95.91 & &79.40 &97.86 &\cellcolor[HTML]{68c08c}88.63 & &86.89 &90.59 &\cellcolor[HTML]{a0ce91}88.74 & &89.09 &97.72 &\cellcolor[HTML]{b4d393}93.41 & &76.28 &93.48 &\cellcolor[HTML]{bbd493}84.88 \\ \midrule

\multicolumn{24}{c}{ViT-L-14 $\downarrow$} \\ \midrule

MCM~\cite{mcm_clip_2022} & & &MSP ($\tau=1$) & &81.29 &94.50 &\cellcolor[HTML]{fce599}87.89 & &65.87 &92.78 &\cellcolor[HTML]{f6e399}79.32 & &80.09 &84.50 &\cellcolor[HTML]{fcdc95}82.30 & &87.20 &97.21 &\cellcolor[HTML]{f2e298}92.21 & &73.28 &91.45 &\cellcolor[HTML]{fee498}82.36 \\
GL-MCM~\cite{gl_mcm_2025} & & &GL-MCM & &79.89 &94.99 &\cellcolor[HTML]{fee398}87.44 & &66.27 &89.91 &\cellcolor[HTML]{fdde96}78.09 & &82.97 &87.97 &\cellcolor[HTML]{dcdd96}85.47 & &87.60 &97.15 &\cellcolor[HTML]{e9e098}92.38 & &73.94 &90.85 &\cellcolor[HTML]{ffe599}82.39 \\
TAG~\cite{tag_2024}& & &MSP +TAG & &80.16 &93.83 &\cellcolor[HTML]{fddf96}87.00 & &69.35 &90.90 &\cellcolor[HTML]{eae098}80.12 & &77.62 &83.36 &\cellcolor[HTML]{f9cf91}80.49 & &89.43 &97.09 &\cellcolor[HTML]{bbd493}93.26 & &75.72 &90.70 &\cellcolor[HTML]{e9e097}83.21 \\

TAG~\cite{tag_2024}& & &MaxLogit+TAG & &84.82 &97.98 &\cellcolor[HTML]{c0d694}91.40 & &64.84 &83.17 &\cellcolor[HTML]{f3b286}74.01 & &78.53 &88.87 &\cellcolor[HTML]{fde599}83.70 & &86.09 &95.32 &\cellcolor[HTML]{f9cf91}90.71 & &71.73 &88.29 &\cellcolor[HTML]{f7c68d}80.01 \\

NegLabel~\cite{neglabel-2024} & &\tick &NegLabel & &74.15 &98.80 &\cellcolor[HTML]{fcda95}86.48 & &63.69 &91.67 &\cellcolor[HTML]{fcd995}77.68 & &72.14 &67.32 &\cellcolor[HTML]{e88476}69.73 & &88.81 &98.13 &\cellcolor[HTML]{b0d292}93.47 & &77.89 &94.77 &\cellcolor[HTML]{93ca90}86.33 \\

\rowcolor{Gray}\textbf{ours} &\textbf{} & &[cls+clip] & &91.63 &98.09 &\cellcolor[HTML]{85c78f}94.86 & &74.50 &96.59 &\cellcolor[HTML]{97cb90}85.54 & &90.64 &92.98 &\cellcolor[HTML]{67bf8c}91.81 & &90.65 &97.73 &\cellcolor[HTML]{8bc88f}94.19 & &80.34 &93.84 &\cellcolor[HTML]{7ec58e}87.09 \\

\rowcolor{Gray}\textbf{ours} &\multirow{-2}{*}{R-50} & &[cls+clip+probe] & &95.29 &99.68 &\cellcolor[HTML]{58bc8b}97.48 & &80.93 &98.50 &\cellcolor[HTML]{57bb8a}89.72 & &91.62 &93.64 &\cellcolor[HTML]{57bb8a}92.63 & &91.87 &98.47 &\cellcolor[HTML]{57bb8a}95.17 & &81.73 &95.23 &\cellcolor[HTML]{57bb8a}88.48 \\

\bottomrule
\end{tabular}%
}

\caption{Comparison with state-of-the-art CLIP-based OOD detection methods, grouped by the CLIP architecture. All scores are AUROC in \%. Average OOD scores are color-coded from worst (red) to best (green).}

\label{tab:comparison-clipbbased}
\end{table*}

\begin{table*}[!htp]\centering

\resizebox{\textwidth}{!}{%
\begin{tabular}{lrrrrrrrrrrrrrrrrrrrrrr}\toprule
&ID dataset $\rightarrow$ &\multicolumn{4}{c}{CIFAR-100} & &\multicolumn{7}{c}{ImageNet-200} & &\multicolumn{7}{c}{ImageNet-1K} \\
&classifier $\rightarrow$ &\multicolumn{4}{c}{R-18 (scratch)} & &\multicolumn{7}{c}{R-18 (scratch)} & &\multicolumn{7}{c}{R-50} \\\cmidrule{2-6}\cmidrule{8-14}\cmidrule{16-22}

&eval setting $\rightarrow$ &\multicolumn{4}{c}{standard} & &\multicolumn{4}{c}{standard} & & \multicolumn{2}{c}{\textbf{full-spectrum}} & & \multicolumn{4}{c}{standard} & &\multicolumn{2}{c}{\textbf{full-spectrum}} \\\cmidrule{2-6}\cmidrule{8-11} \cmidrule{13-14} \cmidrule{16-19} \cmidrule{21-22}

\makecell{extra model} &OOD detector &\makecell{near\\OOD} &\makecell{far\\OOD} &\makecell{avg\\OOD} &\makecell{ID\\acc.} & &\makecell{near\\OOD} &\makecell{far\\OOD} &\makecell{avg\\OOD} &\makecell{ID\\acc.} & &\makecell{avg\\OOD}  &\makecell{ID\\acc.} & &\makecell{near\\OOD} &\makecell{far\\OOD} &\makecell{avg\\OOD} &\makecell{ID\\acc.} & &\makecell{avg\\OOD}  &\makecell{ID\\acc.} \\\midrule
&MSP~\cite{msp-2017} &80.27 &77.76 &\cellcolor[HTML]{eb947b}79.02 &77.25 & &83.34 &90.13 &\cellcolor[HTML]{f6bf8b}86.73 &86.37& &\cellcolor[HTML]{f2ae85}60.23 &43.92 & &76.02 &85.23 &\cellcolor[HTML]{e67c73}80.63 &76.18& &\cellcolor[HTML]{f4ba89}66.55 &54.35 \\

&TempScale~\cite{tempscaling-2017} &80.90 &78.74 &\cellcolor[HTML]{f2b186}79.82 &77.25 & &83.69 &90.82 &\cellcolor[HTML]{fad191}87.25 &86.37& &\cellcolor[HTML]{efa280}59.75 &43.92 & &77.14 &87.56 &\cellcolor[HTML]{f0a782}82.35 &76.18& &\cellcolor[HTML]{f6c08b}66.83 &54.35 \\

&RMDS~\cite{rmds-2021} &80.15 &82.92 &\cellcolor[HTML]{fbe499}81.53 &77.25 & &82.57 &88.06 &\cellcolor[HTML]{ea8e79}85.31 &86.37& &\cellcolor[HTML]{fde599}62.61 &43.92 & &76.99 &86.38 &\cellcolor[HTML]{ec967c}81.69 &76.18& &\cellcolor[HTML]{fbd493}67.68 &54.35 \\
&KNN~\cite{knn-2022} &80.18 &82.40 &\cellcolor[HTML]{ffe599}81.29 &77.25 & &81.57 &93.16 &\cellcolor[HTML]{fbd593}87.36 &86.37& &\cellcolor[HTML]{f4b788}60.56 &43.92 & &71.10 &90.18 &\cellcolor[HTML]{e67c73}80.64 &76.18& &\cellcolor[HTML]{e67c73}63.92 &54.35 \\
&MaxLogit~\cite{mls-2022} &81.05 &79.67 &\cellcolor[HTML]{f7c48d}80.36 &77.25 & &82.90 &91.11 &\cellcolor[HTML]{f8c88e}87.00 &86.37& &\cellcolor[HTML]{e67c73}58.27 &43.92 & &76.46 &89.57 &\cellcolor[HTML]{f4b788}83.01 &76.18& &\cellcolor[HTML]{efa582}65.70 &54.35 \\
&VIM~\cite{vim-2022} &74.98 &81.70 &\cellcolor[HTML]{e67c73}78.34 &77.25 & &78.68 &91.26 &\cellcolor[HTML]{e78275}84.97 &86.37& &\cellcolor[HTML]{fee297}62.20 &43.92 & &72.08 &92.68 &\cellcolor[HTML]{f0a782}82.38 &76.18& &\cellcolor[HTML]{f3b487}66.30 &54.35 \\

&ASH~\cite{ash-2023} &78.20 &80.58 &\cellcolor[HTML]{efa180}79.39 &77.25 & &82.38 &93.90 &\cellcolor[HTML]{f8e499}88.14 &86.37& &\cellcolor[HTML]{e5df97}65.11 &43.92 & &78.17 &95.74 &\cellcolor[HTML]{d2da95}86.95 &76.18& &\cellcolor[HTML]{96cb90}73.64 &54.35 \\
&GEN~\cite{gen-ood-posthoc-2023} &81.31 &79.68 &\cellcolor[HTML]{f8c98f}80.50 &77.25 & &83.68 &91.36 &\cellcolor[HTML]{fcda95}87.52 &86.37& &\cellcolor[HTML]{e78275}58.52 &43.92 & &76.85 &89.76 &\cellcolor[HTML]{f5bf8b}83.31 &76.18& &\cellcolor[HTML]{f2ae85}66.08 &54.35 \\

&SCALE~\cite{scale-2024} &80.99 &81.42 &\cellcolor[HTML]{fee398}81.20 &77.26 & &84.84 &93.98 &\cellcolor[HTML]{dadc96}89.41 &86.37& &\cellcolor[HTML]{e1de97}65.55 &43.92 & &81.36 &96.53 &\cellcolor[HTML]{a8d092}88.94 &76.18& &\cellcolor[HTML]{57bb8a}76.76 &54.35 \\
&CombOOD~\cite{combood-2024} &78.77 &85.87 &\cellcolor[HTML]{eee198}82.32 &77.25 & &95.74 &92.57 &\cellcolor[HTML]{69c08c}94.16 &86.37& &- &43.92 & &95.22 &90.24 &\cellcolor[HTML]{57bb8a}92.73 &76.18& &- &- \\
&NAC~\cite{nac-2024} &75.90 &86.98 &\cellcolor[HTML]{fce599}81.44 &77.26 & &78.52 &91.02 &\cellcolor[HTML]{e67c73}84.77 &86.37& &\cellcolor[HTML]{e67d73}58.31 &43.92 & &74.43 &95.29 &\cellcolor[HTML]{ffe599}84.86 &76.19& &\cellcolor[HTML]{99cc90}73.48 &54.35 \\
&fDBD~\cite{fdbd-2024} &81.26 &79.85 &\cellcolor[HTML]{f9cb8f}80.56 &77.18 & &84.27 &93.45 &\cellcolor[HTML]{e7df97}88.86 &86.37& &\cellcolor[HTML]{ffe599}62.32 &43.92 & &- &- &- &- & &- &- \\
&Weiper+KLD~\cite{weiperkld-2024} &81.37 &79.01 &\cellcolor[HTML]{f5be8b}80.19 &77.25 & &- &- &- &- & &- &- & &80.05 &95.54 &\cellcolor[HTML]{c0d694}87.80 &76.18& &- &- \\

\midrule
\rowcolor{Gray} &[cls+clip] &84.50 &87.60 &\cellcolor[HTML]{b0d292}86.05 &81.92 & &87.78 &96.79 &\cellcolor[HTML]{96cb90}92.28 &93.18 & & \cellcolor[HTML]{a1ce91}72.36 &75.42 & &76.72 &92.86 &\cellcolor[HTML]{fee498}84.79 &78.81 &&\cellcolor[HTML]{cfd995}70.77 &64.68 \\
\rowcolor{Gray}\multirow{-2}{*}{CLIP B-16} &[cls+clip+probe] &85.37 &89.51 &\cellcolor[HTML]{9acc90}87.44 &83.94 & &88.83 &97.54 &\cellcolor[HTML]{80c68e}93.19 &94.47 &&\cellcolor[HTML]{a5cf91}71.92 &78.08 & &76.28 &93.48 &\cellcolor[HTML]{fee599}84.88 &79.76& &\cellcolor[HTML]{f2e298}69.05 &66.17 \\
\rowcolor{Gray} &[cls+clip] &87.84 &90.19 &\cellcolor[HTML]{80c68e}89.02 &84.99 & &80.34 &93.84 &\cellcolor[HTML]{76c38d}93.63 &94.86& &\cellcolor[HTML]{6dc18c}77.84 &85.09 & &80.34 &93.84 &\cellcolor[HTML]{cfd995}87.09 &80.79& &\cellcolor[HTML]{73c28d}75.40 &71.37 \\
\rowcolor{Gray}\multirow{-2}{*}{CLIP L-14} &[cls+clip+probe] &90.07 &92.84 &\cellcolor[HTML]{57bb8a}91.45 &87.92 & &81.73 &95.23 &\cellcolor[HTML]{57bb8a}94.90 &96.33& &\cellcolor[HTML]{57bb8a}80.11 &88.07 & &81.73 &95.23 &\cellcolor[HTML]{b2d293}88.48 &82.99& &\cellcolor[HTML]{72c28d}75.44 &74.33 \\
\bottomrule

\end{tabular}%
}
\caption{Comparison with state-of-the-art classifier-based OOD detection methods. Baseline results are directly taken from the OpenOOD leaderboard (a few are missing) and are sorted by publication year. All scores are in \%.}
\label{tab:comparison-openood}
\end{table*}

\subsection{Comparison to classifier-based OOD detection}

We compare our approach to post-hoc methods from the OpenOOD leaderboard. Specifically, we consider the top-5 methods in near OOD and far OOD for each of the three ID datasets in this benchmark CIFAR100~\cite{cifar-dataset-2009}, ImageNet200 and ImageNet1K~\cite{imagenet-2009}, including the \textit{full-spectrum} setting (where $D_{test}$ contains covariate-shifted ID images). We also include MSP~\cite{msp-2017} for reference. Results are in Table~\ref{tab:comparison-openood}. We include the [cls+clip] variant in the comparison which does not need any training beyond the base classifier.

First, note that there is a significant gain in classification accuracy with our approach, especially in the full-spectrum setting: 31-44\% on ImageNet-200 and 10-20\% on ImageNet-1K (depending on the CLIP model and ensemble). In OOD detection our method outperforms most approaches, except on the ImageNet-1K benchmark where SCALE~\cite{scale-2024} and CombOOD~\cite{combood-2024} have the advantage. 

\begin{figure}[t]
    \centering
     \includegraphics[width=1\linewidth]{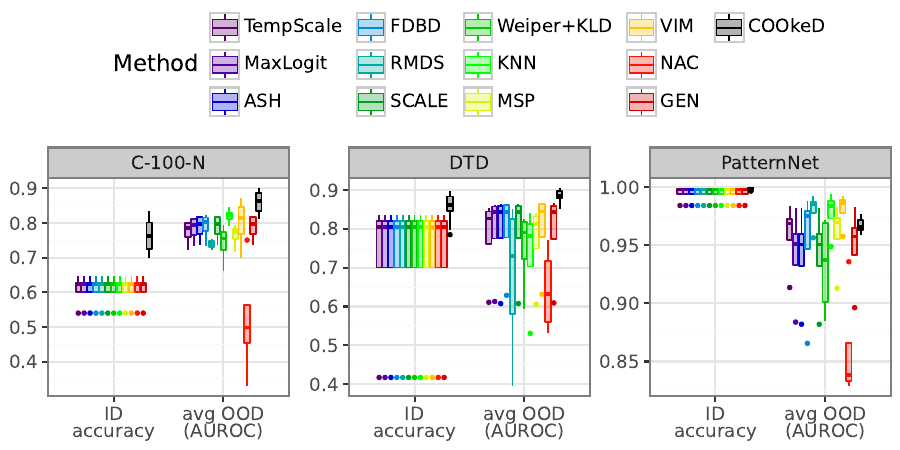}
    \caption{Classification and OOD detection performance beyond the OpenOOD benchmark. Note that baseline methods have the same accuracy since they share the same underlying classifier. Boxplot distribution is across 4 classifier variants and 4 CLIP variants (for COOkeD only).}
    \label{fig:openood-vs-cooked}
\end{figure}

In Figure~\ref{fig:openood-vs-cooked}, we compare COOkeD to the same set of baseline methods (except CombOOD~\cite{combood-2024} due to implementation issues) beyond the OpenOOD benchmark, on a wider range of classifiers trained on different ID datasets. None of the methods consistently outperform COOkeD, and they are inherently limited in their classification accuracy.

\section{Discussion}

\textbf{A hungry method?} Our aim is to show how OOD detection can be pushed by making full use of what is already there (plenty of pre-trained CLIP models), while acknowledging that dataset-specific training is necessary in many domains where CLIP's zero-shot performance is insufficient. Using our full method requires two trained classifiers (standard and linear probe) and also using CLIP at test-time alongside a separate model (e.g. ResNet). This is not minimalistic by any means, and is not well-suited for applications with limited computational resources at test-time. However, it does provide a very simple but robust and practical solution to OOD detection while pushing classification accuracy. In comparison to deep ensembling~\cite{deepensembles-2017}, COOkeD is a \textit{cheap} ensemble, as it only requires training one full model and a linear classifier (rather than multiple full models), and only costs two forward passes at test-time. COOkeD is also cheaper than some CLIP-based approaches which require training a text encoder and using it at test-time~\cite{zeroshot-ood-clip-2022,clipn-2023}.

\textbf{What next?}  We hope that this opens the door for future work exploring the complementarity of zero-shot and supervised classifiers. COOkeD is a strong baseline made up of simple, standard building blocks: a standard CE classifier, a popular CLIP variant, a standard CE linear probe, default CLIP prompts \& temperature, equal weights in the ensemble, and a basic OOD scoring function. The fact that robust performance is achieved with no ``bells and whistles'' is encouraging and makes COOkeD broadly applicable. This leaves plenty of room for improvement along many axes, drawing from recent advances in both the full-shot and the zero-shot OOD detection literature.

\section*{Acknowledgements}

Galadrielle Humblot-Renaux was supported by the Danish Data Science Academy, which is funded by the Novo Nordisk Foundation (NNF21SA0069429) and VILLUM FONDEN (40516). Thanks also to the Pioneer Centre for AI, DNRF grant number P1. 

This work has been partially supported by the Spanish project PID2022-136436NB-I00 and by ICREA under the ICREA Academia programme.

This project was provided with computer and storage resources by GENCI at IDRIS thanks to the grant 2025-104306 on the supercomputer Jean Zay.

{
    \small
    \bibliographystyle{ieeenat_fullname}
    \bibliography{refs}
}

\clearpage

\onecolumn
\appendix
\renewcommand{\thesection}{\Alph{section}}
\renewcommand{\thesubsection}{\Alph{section}.\arabic{subsection}}
\section{Approach}\label{supp:approach}

\subsection{Code snippet}

\begin{python}

clip.eval() # CLIP model from open_clip
probe.eval() # fully connected layer trained to classify (un-normalized) CLIP image features
classifier.eval() # e.g. Resnet50 trained to classify images

# note: different normalization for CLIP image encoder vs. standard classifier
image_normalized_clip = normalize_for_clip(image)
image_normalized_cls = normalize_for_cls(image)

# 1. get zero-shot CLIP prediction
prompt_template = "a photo of a [cls]"
prompts = [prompt_template.replace("[cls]",f"{class_mapping[idx]}") for idx in range(num_id_classes)]

prompt_tokenized = clip_tokenizer(prompts)
prompt_features = clip.encode_text(prompt_tokenized)
prompt_features_normed = prompt_features / prompt_features.norm(dim=-1, keepdim=True)

clip_image_features = clip.encode_image(image_normalized_clip)
clip_image_features_normed = image_features / image_features.norm(dim=-1, keepdim=True)
text_sim = (image_features_normed @ prompt_features_normed.T)
            
logits_clip_t100 = 100 * text_sim
softmax_clip_t100 = logits_clip_t100.softmax(dim=1)

# 2. get probe CLIP prediction
logits_probe = probe(clip_image_features)
softmax_probe = logits_probe.softmax(dim=1)

# 3. get classifier prediction
logits_classifier = classifier(image_normalized_cls)
softmax_classifier = logits_classifier.softmax(dim=1)

# 4. combined prediction
softmax_ensemble = torch.stack([softmax_clip_t100, softmax_probe, softmax_classifier]).mean(0)

# class prediction and OOD scores
pred = softmax_ensemble.argmax(dim=1)

msp = softmax_ensemble.max(dim=1)
entropy = torch.distributions.Categorical(probs=softmax_ensemble).entropy()
                
\end{python}

\section{Experimental set-up}\label{supp:xp-setup}

\subsection{Datasets}

\subsubsection{OpenOOD}

For a given ID dataset (e.g. CIFAR100), OpenOOD~\cite{openood_1.5_2023} defines a ID\textsubscript{train}, ID\textsubscript{val}, ID\textsubscript{test} sets and multiple OOD\textsubscript{test} sets, and we use these without modification. ID\textsubscript{train} is used for training the classifier and certain OOD detectors (e.g. KNN, RMDS) use it to compute dataset statistics. ID\textsubscript{val} is used for early stopping and for OOD hyperparameter tuning by certain methods (e.g. TempScale). ID\textsubscript{test} and OOD\textsubscript{test} are used for evaluating classification and OOD detection performance.

\noindent\textbf{ID datasets}:
\begin{itemize}
    \item CIFAR100 needs no introduction. ID\textsubscript{train} is the official training set. ID\textsubscript{val} (1000 images) and ID\textsubscript{test} (9000 images) are subsets of the official test set. 
    \item CIFAR100-N~\cite{cifar-noisy-dataset-2022} contains the same images as CIFAR100 but provides a re-annotated version of CIFAR100's ID\textsubscript{train} set with real label noise ($\approx$ 40\% noise rate arising from crowdsourced annotations). The ID\textsubscript{val} and ID\textsubscript{test} are cleanly labelled and identical to those in CIFAR100. Note that this dataset is not part of the OpenOOD benchmark/leaderboard but we apply identical splits and pre-processing.
    \item ImageNet-1K corresponds to the ILSVRC2012 version of ImageNet. ID\textsubscript{train} is the official training set. ID\textsubscript{val} (5000 images) and ID\textsubscript{test} (45000 images) are subsets of the official validation set. 
    \item ImageNet-200 is a 200-class subset of ImageNet-1K defined in~\cite{imagenet-r-2021}. ID\textsubscript{train} is a subset of the official ImageNet-1K training set. ID\textsubscript{val} (1000 images) and ID\textsubscript{test} (9000 images) are subsets of the official ImageNet-1K validation set. 
    \item In the covariate shifted versions of ImageNet-1K and ImageNet-200, the original ID\textsubscript{test} is concatenated with images from ImageNet-R~\cite{imagenet_r-2021}, ImageNet-V2~\cite{imagenet_v2} and ImageNet-C~\cite{imagenet_c-2019}. We use the exact same sets as those defined in the OpenOOD full-spectrum benchmark.
\end{itemize}

\noindent Note that our method does not make use of ID\textsubscript{val}, as we use the last training checkpoint for evaluation, and involves no OOD hyper-parameters.

\subsubsection{OODDB}

OODDB~\cite{ooddb} provides 5 open-set splits for several datasets. We use first split (split 0) for PatternNet and DTD which defines the following set of ID vs OOD classes:

\noindent\textbf{PatternNet}:
\begin{itemize}
    \item ID: {0: 'oil well', 1: 'overpass', 2: 'railway', 3: 'basketball court', 4: 'river', 5: 'wastewater treatment plant', 6: 'christmas tree farm', 7: 'sparse residential', 8: 'chaparral', 9: 'solar panel', 10: 'parking lot', 11: 'airplane', 12: 'golf course', 13: 'bridge', 14: 'freeway', 15: 'transformer station', 16: 'mobile home park', 17: 'nursing home', 18: 'beach'}
    \item OOD: {19: 'ferry terminal', 20: 'storage tank', 21: 'forest', 22: 'coastal mansion', 23: 'swimming pool', 24: 'closed road', 25: 'shipping yard', 26: 'dense residential', 27: 'runway', 28: 'tennis court', 29: 'crosswalk', 30: 'intersection', 31: 'runway marking', 32: 'cemetery', 33: 'baseball field', 34: 'oil gas field', 35: 'parking space', 36: 'football field', 37: 'harbor'}
\end{itemize}

\noindent\textbf{DTD}:
\begin{itemize}
    \item ID: {0: 'crystalline', 1: 'fibrous', 2: 'spiralled', 3: 'stratified', 4: 'matted', 5: 'porous', 6: 'scaly', 7: 'stained', 8: 'pleated', 9: 'flecked', 10: 'pitted', 11: 'meshed', 12: 'freckled', 13: 'waffled', 14: 'cracked', 15: 'potholed', 16: 'cobwebbed', 17: 'swirly', 18: 'polka-dotted', 19: 'striped', 20: 'studded', 21: 'grooved', 22: 'woven'}
    \item OOD: {23: 'banded', 24: 'blotchy', 25: 'braided', 26: 'bubbly', 27: 'bumpy', 28: 'chequered', 29: 'crosshatched', 30: 'dotted', 31: 'frilly', 32: 'gauzy', 33: 'grid', 34: 'honeycombed', 35: 'interlaced', 36: 'knitted', 37: 'lacelike', 38: 'lined', 39: 'marbled', 40: 'paisley', 41: 'perforated', 42: 'smeared', 43: 'sprinkled', 44: 'veined', 45: 'wrinkled', 46: 'zigzagged'}
\end{itemize}

\subsection{Models}

\subsubsection{Standard classifiers}

For standard ImageNet1K classifiers, we use pretrained models from torchvision. Specifically:
ResNet18: \verb|resnet18-f37072fd.pth|, ResNet50: \verb|resnet50-0676ba61.pth|, ViT-B-16: \verb|vit_b_16-c867db91.pth|.

For the other ID datasets (CIFAR100, PatternNet and DTD) we train a ResNet18, ResNet50 and ViT-B-16 with the following optimizer and scheduler:
\begin{python}
optimizer = torch.optim.SGD(
    net.parameters(), lr, momentum=0.9, weight_decay=0.0005, nesterov=True,
)

def cosine_annealing(step, total_steps, lr_max, lr_min):
    return lr_min + (lr_max - lr_min) * 0.5 * (1 + np.cos(step / total_steps * np.pi))

scheduler = torch.optim.lr_scheduler.LambdaLR(
    self.optimizer, lr_lambda=lambda step: cosine_annealing( step, num_epochs * len(train_loader), 1, 1e-6 / lr )
)
\end{python}
for 100 epochs with a learning rate of 0.1 when training from scratch and 0.01 when fine-tuning from a torchvision checkpoint. The batch size is 128 for CIFAR datasets otherwise 256.

\subsubsection{Zero-shot CLIP and Probe CLIP}

We use the following CLIP models in our experiments:
\begin{python}
    open_clip.create_model_and_transforms("ViT-B-32", pretrained="openai")
    open_clip.create_model_and_transforms("ViT-B-16", pretrained="openai")
    open_clip.create_model_and_transforms("ViT-L-14", pretrained="openai")
    open_clip.create_model_and_transforms("ViT-H-14", pretrained="laion2b_s32b_b79k")
\end{python}

The linear probe is implemented as a single linear/fully connected layer with bias, taking as input the un-normalized image embeddings extracted by CLIP (pre-extracted before training). It is trained for 20 epochs with a learning rate of 0.1, and the same optimizer, scheduler and batch size as for the standard classifier (see above).

\subsection{Image pre-processing}


To summarize (see below for more details): an input size of $224\times224$ is used for all datasets. At test-time, images are resized such that the shortest side is $224$ and then center-cropped to $224\times224$.
\begin{itemize}
    \item Classifier model: input images are normalized based on dataset-specific mean and std. During training, only basic data augmentations from OpenOOD are applied (resize, flip, random crop for CIFAR100, and flip, random resized crop for the rest).
    \item CLIP models: input images are normalized based on the OpenAI mean and std. The linear classifier is trained on un-normalized CLIP image embeddings which were extracted and stored before training. Thus, no augmentations are applied. 
\end{itemize}

\vspace{0.5em}

On the standard classifier side, we use the same pre-processing procedure as the OpenOOD codebase. On the CLIP side, we align with previous work on CLIP-based OOD detection.

\subsubsection{Image normalization}

\noindent For standard classifiers, the following normalization paramaters are used:
\begin{itemize}
    \item CIFAR100: {\footnotesize\verb|mean=(0.5071, 0.4867, 0.4408), std=(0.2675, 0.2565, 0.2761)|}
    \item ImageNet-200 and ImageNet-1k: {\footnotesize\verb|mean=(0.485, 0.456, 0.406), (0.229, 0.224, 0.225)|}
    \item PatternNet: {\footnotesize\verb|mean=(0.5, 0.5, 0.5), std=(0.5, 0.5, 0.5)|}
    \item DTD: {\footnotesize\verb|mean=(0.5, 0.5, 0.5), std=(0.5, 0.5, 0.5)|}
\end{itemize}

\vspace{0.5em}

\noindent For CLIP (zero-shot and probe), the same normalization parameters are used across all datasets: {\footnotesize\verb|mean=(0.48145466, 0.4578275, 0.40821073), std=(0.26862954, 0.26130258, 0.27577711)|}

\subsection{Image transformations}

For standard classifiers, the following image transformations are used:
\begin{python}
# training CIFAR100
tvs_trans.Compose([
        tvs_trans.Resize(224),
        tvs_trans.RandomHorizontalFlip(),
        tvs_trans.RandomCrop(image_size, padding=4),
        tvs_trans.ToTensor(),
        normalize_cls
    ])

# training other datasets
tvs_trans.Compose([
        tvs_trans.RandomHorizontalFlip(),
        tvs_trans.RandomResizedCrop(image_size),
        tvs_trans.ToTensor(),
        normalize_cls
    ])

# evaluation
tvs_trans.Compose([
        tvs_trans.Resize(224),
        tvs_trans.CenterCrop(224),
        tvs_trans.ToTensor(),
        normalize_cls
    ])
\end{python}

\noindent For CLIP (zero-shot and probe), the following image transformations are used:

\begin{python}
tvs_trans.Compose([
        tvs_trans.Resize(224),
        tvs_trans.CenterCrop(224),
        tvs_trans.ToTensor(),
        normalize_clip
    ])
\end{python}

\subsection{Class names}

Class names are used when constructing prompts for zero-shot CLIP.

\noindent\textbf{ImageNet-1K and ImageNet-200} Similarly to existing work in zero-shot CLIP OOD detection~\cite{mcm_clip_2022,gl_mcm_2025,tag_2024,neglabel-2024}, we use the ImageNet class names provided in the OpenAI CLIP repository, which are ``prompt-engineered" and deviate from the raw class names. For example, for class 884, the class name "vaulted or arched ceiling" is used instead of the raw/original class name "vault". Note that using raw vs. prompt-engineered class names for constructing CLIP text prompts has a non-negligible impact on zero-shot performance. 

\noindent\textbf{CIFAR100}  We use the raw class names, replacing underscores with spaces.

\noindent\textbf{OODDB PatternNet \& OODDB DTD} We use the raw class names, exactly as listed above.

\subsection{Evaluation metrics}

\begin{itemize}
    \item The classification accuracy is computed on ID\textsubscript{test}.
    \item The OOD AUROC and OOD FPR@95 are computed on the ID test set vs. a specific OOD dataset. ID samples are treated as the positive label.
\end{itemize}

\section{Baselines}\label{supp:baselines}

\subsection{CLIP-based approaches}

When evaluating previous CLIP-based methods, we used their official implementation for the model, temperature, prompt template and image pre-processing but replaced the data splits with those described in \ref{sec:setup-datasets}. ZOC~\cite{zeroshot-ood-clip-2022} generates candidate negative prompts using a trained BERT encoder which can be used across multiple ID datasets, therefore we use the model weights provided by the authors. NegLabel~\cite{neglabel-2024} generates candidate negative prompts based on the ID class names - we therefore obtain a different set of negative prompts for each ID dataset. 

\subsection{Classifier-based approaches}

All methods in our comparison except NAC~\cite{nac-2024}, Weiper-KLD~\cite{weiperkld-2024} and CombOOD~\cite{combood-2024} are integrated in the OpenOOD codebase, and we apply them with their default configuration. For Weiper and NAC, we use the official implementation \url{https://github.com/mgranz/weiper} and \url{https://github.com/BierOne/ood_coverage/} with the suggested configuration provided for each architecture. We had issues reproducing CombOOD results from its official implementation \url{https://github.com/rmagesh148/combood/} and therefore did not include it in the final experiment.

\section{Results}\label{supp:results}

\subsection{COOkeD in action}

\begin{figure*}[h]
    \begin{subfigure}{\linewidth}
        \includegraphics[width=\linewidth]{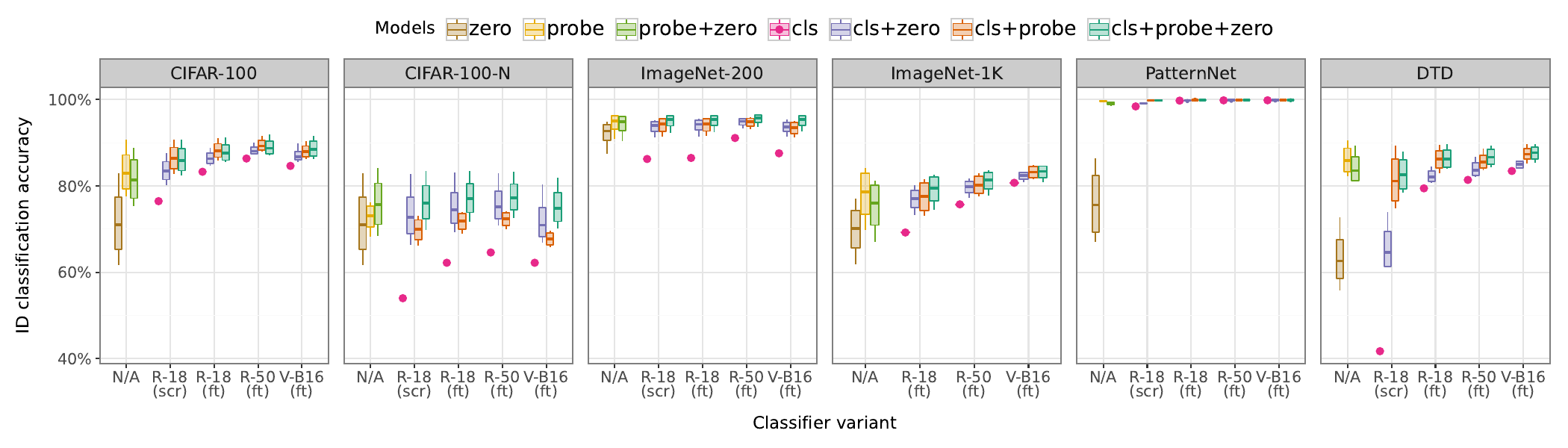}
    \end{subfigure}
    \caption{Classification accuracy with different combinations of Zero-shot / Probe / Standard classifier models. Boxplots show the distribution across different CLIP and classifier variants.}
    \label{fig:supp_acc_ensemble_combinations}
\end{figure*}

Figures~\ref{fig:supp_acc_ensemble_combinations} (classification) and ~\ref{fig:supp_acc_ensemble_combinations} (OOD detection) extend Figure~\ref{fig:ood_ensemble_combinations} from the main text by showing performance on a per-dataset basis. Figure~\ref{fig:supp_acc_ensemble_combinations} also provides results in terms of FPR95 and with the Entropy vs. MSP as scoring function.

\begin{figure*}
    \begin{subfigure}{\linewidth}
        \includegraphics[width=\linewidth]{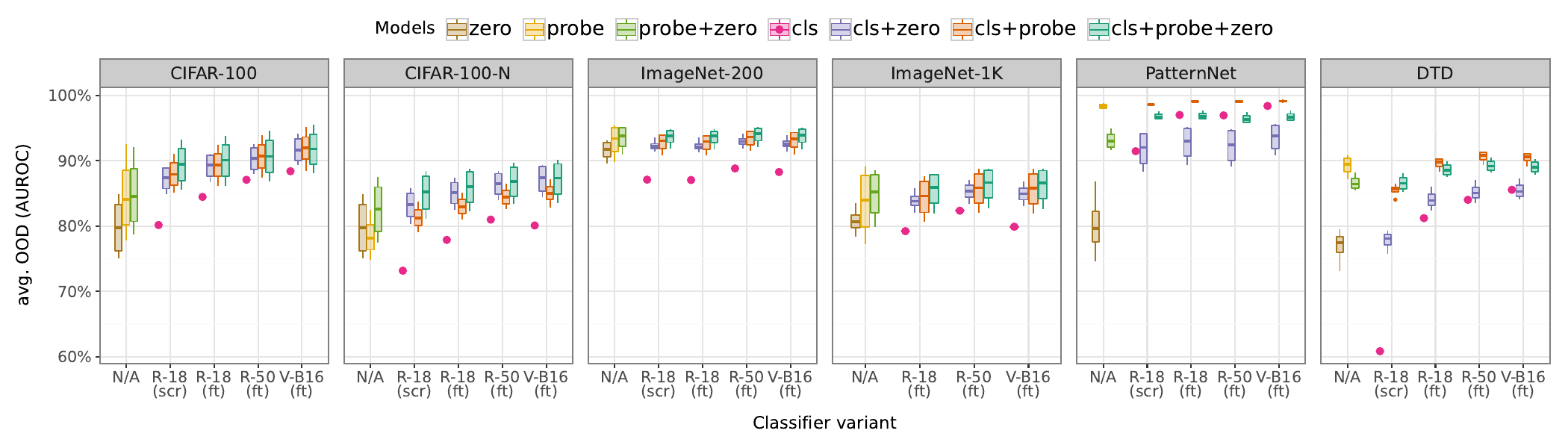}
        \caption{AUROC - Entropy as OOD scoring function}

        \includegraphics[width=\linewidth]{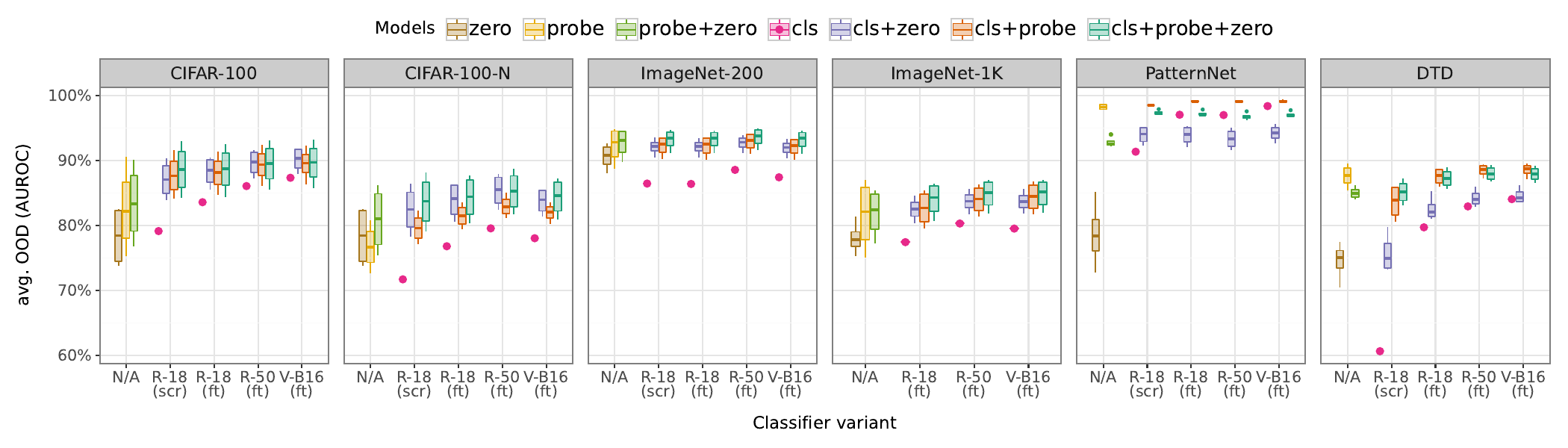}
        \caption{AUROC - MSP as OOD scoring function}

        \includegraphics[width=\linewidth]{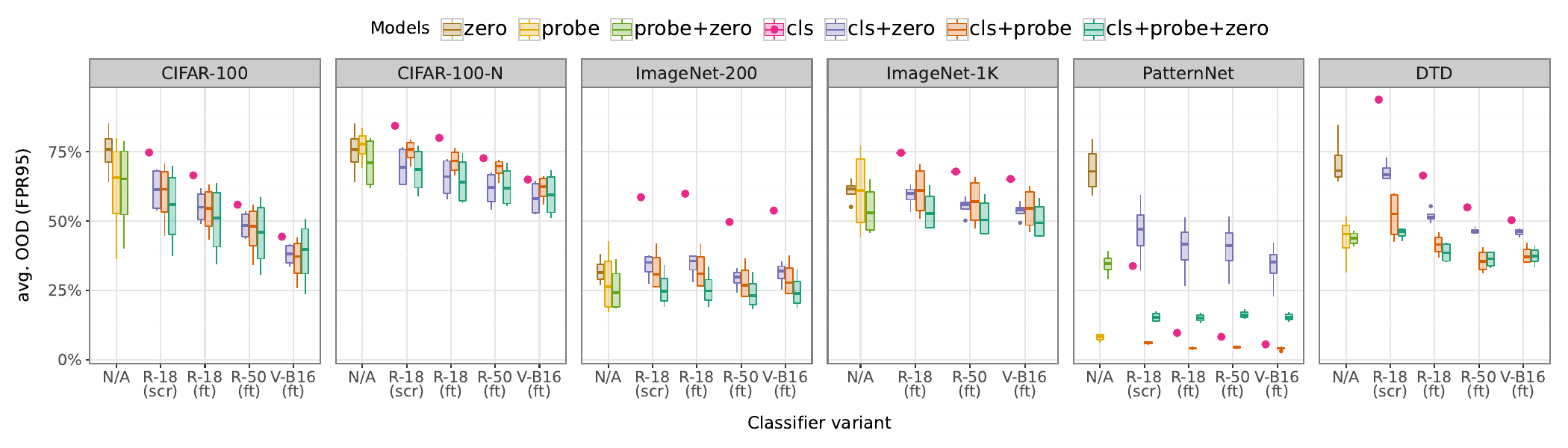}
        \caption{FPR95 - Entropy as OOD scoring function}
        \includegraphics[width=\linewidth]{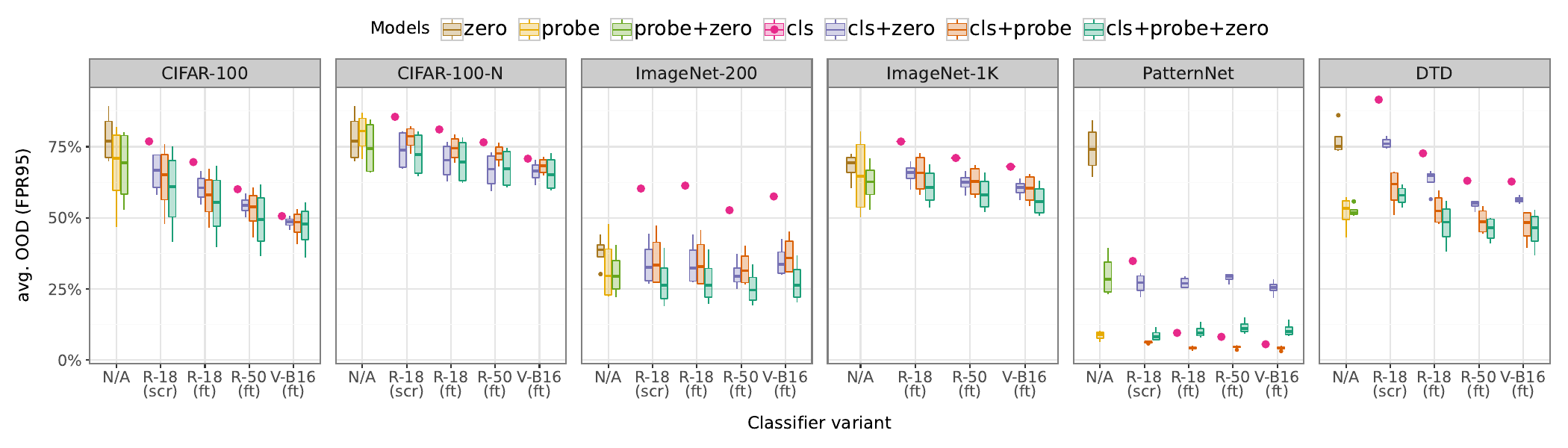}
        \caption{FPR95 - MSP as OOD scoring function}
    \end{subfigure}
    \caption{OOD detection performance with different combinations of Zero-shot / Probe / Standard classifier models. Boxplots show the distribution across different CLIP and classifier variants.}
    \label{fig:supp_ood_ensemble_combinations}
\end{figure*}

\clearpage

\subsection{Comparison to classifier-based OOD detection}

\begin{figure}[h]
    \centering
     \includegraphics[width=1\linewidth]{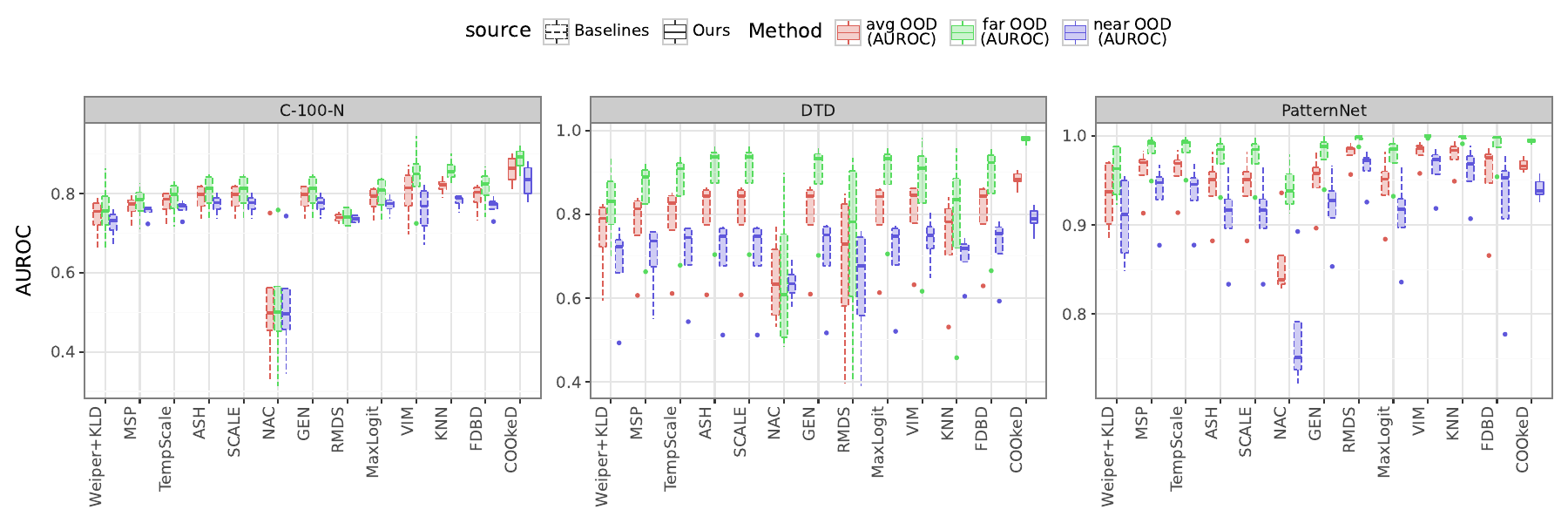}
     \includegraphics[width=1\linewidth]{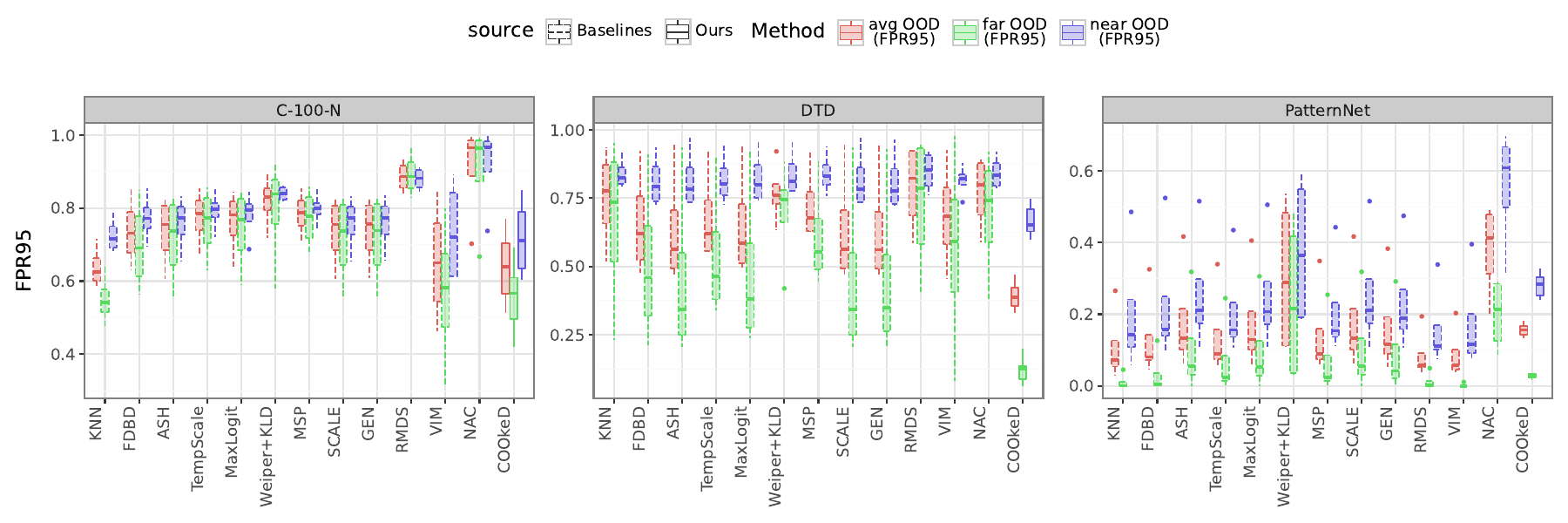}
    \caption{OOD detection performance (AUROC and FPR95) beyond the OpenOOD benchmark. Boxplot distribution is across 4 classifier variants and 4 CLIP variants (for COOkeD only).}
    \label{fig:supp_openood-vs-cooked}
\end{figure}

\subsection{Results for each OOD dataset}

The four tables below contain the ``raw'' results for each ID/OOD dataset pair, classifier variant and CLIP variant, using COOkeD (ensemble of standard classifier, zero-shot CLIP and probe CLIP, with the entropy as OOD score.) All performance is in \%. Besides the ID accuracy, the OOD detection performance for each OOD dataset corresponds to the AUROC. 

\begin{table}
\centering

\vspace{2em}\resizebox{0.9\linewidth}{!}{%
\begin{tabular}{lllccccccc}
\toprule
Dataset & Classifier & CLIP & ID Acc. & SVHN & Texture & TIN & Places365 & MNIST \\
\midrule
CIFAR-100 & ResNet-18 (scr) & ViT-B-32 & 82.38 & 96.06 & 82.25 & 84.74 & 83.98 & 89.28 \\
CIFAR-100 & ResNet-18 (scr) & ViT-B-16 & 83.94 & 95.39 & 84.35 & 86.97 & 84.80 & 93.48 \\
CIFAR-100 & ResNet-18 (scr) & ViT-L-14 & 87.92 & 97.74 & 88.59 & 90.92 & 89.99 & 95.05 \\
CIFAR-100 & ResNet-18 (scr) & ViT-H-14 & 90.68 & 98.64 & 91.40 & 92.50 & 93.42 & 94.70 \\
CIFAR-100 & ResNet-18 (ft) & ViT-B-32 & 85.50 & 96.64 & 88.24 & 86.61 & 81.77 & 83.78 \\
CIFAR-100 & ResNet-18 (ft) & ViT-B-16 & 86.16 & 96.22 & 89.78 & 88.72 & 82.77 & 89.84 \\
CIFAR-100 & ResNet-18 (ft) & ViT-L-14 & 89.01 & 98.05 & 92.98 & 92.49 & 88.43 & 92.08 \\
CIFAR-100 & ResNet-18 (ft) & ViT-H-14 & 91.26 & 98.74 & 95.27 & 93.98 & 92.30 & 91.06 \\
CIFAR-100 & ResNet-50 (ft) & ViT-B-32 & 87.14 & 97.41 & 88.01 & 87.91 & 83.60 & 85.70 \\
CIFAR-100 & ResNet-50 (ft) & ViT-B-16 & 87.57 & 97.04 & 89.48 & 89.83 & 84.49 & 91.34 \\
CIFAR-100 & ResNet-50 (ft) & ViT-L-14 & 89.94 & 98.56 & 92.93 & 93.46 & 89.90 & 93.18 \\
CIFAR-100 & ResNet-50 (ft) & ViT-H-14 & 91.98 & 98.99 & 95.66 & 95.03 & 93.86 & 92.50 \\
CIFAR-100 & ViT-B16 (ft) & ViT-B-32 & 86.23 & 97.47 & 93.06 & 89.38 & 90.28 & 83.85 \\
CIFAR-100 & ViT-B16 (ft) & ViT-B-16 & 87.12 & 96.89 & 93.80 & 91.17 & 90.77 & 90.53 \\
CIFAR-100 & ViT-B16 (ft) & ViT-L-14 & 89.92 & 98.52 & 95.90 & 94.46 & 94.33 & 92.47 \\
CIFAR-100 & ViT-B16 (ft) & ViT-H-14 & 91.58 & 99.11 & 97.75 & 96.11 & 96.99 & 92.12 \\
\bottomrule
\end{tabular}%
}

\vspace{2em}\resizebox{0.9\linewidth}{!}{%
\begin{tabular}{lllrrrrrr}
\toprule
Dataset & Classifier & CLIP & ID Acc. & SVHN & Texture & TIN & Places365 & MNIST \\
\midrule
CIFAR-100-N & ResNet-18 (scr) & ViT-B-32 & 69.89 & 96.09 & 77.18 & 79.86 & 76.59 & 87.89 \\
CIFAR-100-N & ResNet-18 (scr) & ViT-B-16 & 73.19 & 94.82 & 79.97 & 82.29 & 76.97 & 92.23 \\
CIFAR-100-N & ResNet-18 (scr) & ViT-L-14 & 78.98 & 97.44 & 84.19 & 87.26 & 82.89 & 92.96 \\
CIFAR-100-N & ResNet-18 (scr) & ViT-H-14 & 83.46 & 98.38 & 85.32 & 87.40 & 86.03 & 92.09 \\
CIFAR-100-N & ResNet-18 (ft) & ViT-B-32 & 71.74 & 96.82 & 82.54 & 81.37 & 75.50 & 87.17 \\
CIFAR-100-N & ResNet-18 (ft) & ViT-B-16 & 74.60 & 95.78 & 84.63 & 83.58 & 75.86 & 91.34 \\
CIFAR-100-N & ResNet-18 (ft) & ViT-L-14 & 79.56 & 97.93 & 88.07 & 88.24 & 81.70 & 92.29 \\
CIFAR-100-N & ResNet-18 (ft) & ViT-H-14 & 83.52 & 98.57 & 88.74 & 88.19 & 84.59 & 90.26 \\
CIFAR-100-N & ResNet-50 (ft) & ViT-B-32 & 72.54 & 97.93 & 82.38 & 82.70 & 77.66 & 90.78 \\
CIFAR-100-N & ResNet-50 (ft) & ViT-B-16 & 75.11 & 96.96 & 84.47 & 84.63 & 78.03 & 93.98 \\
CIFAR-100-N & ResNet-50 (ft) & ViT-L-14 & 79.48 & 98.49 & 88.15 & 88.96 & 83.44 & 94.57 \\
CIFAR-100-N & ResNet-50 (ft) & ViT-H-14 & 83.34 & 98.93 & 89.34 & 89.11 & 86.49 & 93.17 \\
CIFAR-100-N & ViT-B16 (ft) & ViT-B-32 & 70.14 & 97.44 & 89.07 & 82.66 & 81.02 & 82.57 \\
CIFAR-100-N & ViT-B16 (ft) & ViT-B-16 & 72.37 & 96.62 & 90.46 & 84.79 & 81.26 & 87.95 \\
CIFAR-100-N & ViT-B16 (ft) & ViT-L-14 & 77.39 & 98.25 & 92.92 & 89.28 & 86.50 & 89.22 \\
CIFAR-100-N & ViT-B16 (ft) & ViT-H-14 & 81.98 & 98.85 & 94.18 & 89.52 & 89.62 & 86.04 \\
\bottomrule
\end{tabular}%
}
\caption{COOkeD performance on CIFAR100 (clean training labels vs noisy training labels setting).}

\end{table}

\begin{table}
\centering
\vspace{2em}\resizebox{0.9\linewidth}{!}{%
\begin{tabular}{lllccccccc}
\toprule
Dataset & Classifier & CLIP & ID Acc. & DTD\textsubscript{OOD} & SVHN & PatternNet\textsubscript{OOD} & TIN & Places365 & MNIST \\
\midrule
DTD & ResNet-18 (scr) & ViT-B-32 & 78.48 & 74.11 & 99.53 & 94.39 & 94.45 & 94.85 & 97.88 \\
DTD & ResNet-18 (scr) & ViT-B-16 & 79.67 & 75.08 & 99.34 & 92.17 & 96.27 & 94.28 & 99.68 \\
DTD & ResNet-18 (scr) & ViT-L-14 & 85.43 & 77.29 & 99.62 & 93.98 & 96.66 & 97.06 & 97.51 \\
DTD & ResNet-18 (scr) & ViT-H-14 & 87.93 & 78.45 & 99.88 & 96.98 & 96.18 & 96.94 & 99.22 \\
DTD & ResNet-18 (ft) & ViT-B-32 & 83.91 & 77.50 & 99.82 & 96.32 & 96.77 & 95.80 & 98.04 \\
DTD & ResNet-18 (ft) & ViT-B-16 & 84.57 & 78.31 & 99.76 & 94.46 & 97.54 & 95.08 & 99.64 \\
DTD & ResNet-18 (ft) & ViT-L-14 & 87.83 & 80.03 & 99.71 & 95.79 & 98.43 & 97.91 & 98.08 \\
DTD & ResNet-18 (ft) & ViT-H-14 & 89.78 & 81.10 & 99.93 & 97.88 & 98.36 & 97.78 & 99.45 \\
DTD & ResNet-50 (ft) & ViT-B-32 & 84.35 & 78.58 & 99.85 & 96.66 & 96.84 & 96.12 & 98.56 \\
DTD & ResNet-50 (ft) & ViT-B-16 & 85.22 & 79.40 & 99.83 & 94.91 & 97.77 & 95.43 & 99.90 \\
DTD & ResNet-50 (ft) & ViT-L-14 & 88.15 & 80.93 & 99.85 & 96.30 & 98.63 & 98.22 & 98.34 \\
DTD & ResNet-50 (ft) & ViT-H-14 & 89.35 & 82.34 & 99.89 & 97.96 & 98.29 & 97.93 & 99.35 \\
DTD & ViT-B16 (ft) & ViT-B-32 & 85.33 & 78.04 & 99.90 & 96.24 & 97.46 & 96.74 & 98.13 \\
DTD & ViT-B16 (ft) & ViT-B-16 & 86.63 & 79.19 & 99.89 & 94.42 & 98.16 & 96.11 & 99.77 \\
DTD & ViT-B16 (ft) & ViT-L-14 & 88.80 & 80.79 & 99.84 & 95.83 & 98.73 & 98.49 & 98.18 \\
DTD & ViT-B16 (ft) & ViT-H-14 & 89.78 & 81.84 & 99.93 & 97.64 & 98.50 & 98.18 & 99.55 \\
\bottomrule
\end{tabular}%
}

\vspace{2em}\resizebox{0.9\linewidth}{!}{%
\begin{tabular}{lllccccccc}
\toprule
Dataset & Classifier & CLIP & ID Acc. & PatternNet\textsubscript{OOD} & SVHN & DTD\textsubscript{OOD} & TIN & Places365 & MNIST \\
\midrule
PatternNet & ResNet-18 (scr) & ViT-B-32 & 99.68 & 93.52 & 99.96 & 98.77 & 99.43 & 97.91 & 99.98 \\
PatternNet & ResNet-18 (scr) & ViT-B-16 & 99.71 & 93.51 & 99.99 & 99.05 & 99.55 & 98.58 & 99.99 \\
PatternNet & ResNet-18 (scr) & ViT-L-14 & 99.89 & 95.56 & 100.00 & 99.13 & 99.75 & 99.13 & 100.00 \\
PatternNet & ResNet-18 (scr) & ViT-H-14 & 99.89 & 94.32 & 100.00 & 98.91 & 99.79 & 99.32 & 100.00 \\
PatternNet & ResNet-18 (ft) & ViT-B-32 & 99.91 & 93.56 & 99.98 & 99.29 & 99.57 & 97.78 & 100.00 \\
PatternNet & ResNet-18 (ft) & ViT-B-16 & 99.86 & 93.64 & 100.00 & 99.29 & 99.63 & 98.70 & 100.00 \\
PatternNet & ResNet-18 (ft) & ViT-L-14 & 99.89 & 95.81 & 99.99 & 99.63 & 99.81 & 99.16 & 99.99 \\
PatternNet & ResNet-18 (ft) & ViT-H-14 & 99.93 & 94.70 & 100.00 & 99.52 & 99.83 & 99.36 & 100.00 \\
PatternNet & ResNet-50 (ft) & ViT-B-32 & 99.89 & 92.68 & 99.98 & 98.74 & 99.58 & 98.38 & 99.98 \\
PatternNet & ResNet-50 (ft) & ViT-B-16 & 99.86 & 92.54 & 99.99 & 98.96 & 99.65 & 98.97 & 99.97 \\
PatternNet & ResNet-50 (ft) & ViT-L-14 & 99.88 & 95.29 & 100.00 & 99.55 & 99.83 & 99.39 & 99.98 \\
PatternNet & ResNet-50 (ft) & ViT-H-14 & 99.96 & 93.98 & 100.00 & 99.16 & 99.82 & 99.44 & 99.99 \\
PatternNet & ViT-B16 (ft) & ViT-B-32 & 99.92 & 93.06 & 99.98 & 99.42 & 99.76 & 98.39 & 100.00 \\
PatternNet & ViT-B16 (ft) & ViT-B-16 & 99.87 & 93.23 & 100.00 & 99.36 & 99.80 & 98.92 & 100.00 \\
PatternNet & ViT-B16 (ft) & ViT-L-14 & 99.92 & 95.62 & 100.00 & 99.73 & 99.90 & 99.39 & 100.00 \\
PatternNet & ViT-B16 (ft) & ViT-H-14 & 99.95 & 94.51 & 100.00 & 99.61 & 99.91 & 99.39 & 100.00 \\
\bottomrule

\end{tabular}%
}
\caption{COOkeD performance on DTD and PatternNet.}

\end{table}

\begin{table}
\centering
\vspace{2em}\resizebox{0.9\linewidth}{!}{%
\begin{tabular}{lllrrrrrr}
\toprule
Dataset & Classifier & CLIP & ID Acc. & iNaturalist & Textures & OpenImage-O & NINCO & SSB-Hard \\
\midrule
IN-200 & ResNet-18 (scr) & ViT-B-32 & 92.38 & 98.36 & 95.58 & 96.30 & 90.60 & 83.35 \\
IN-200 & ResNet-18 (scr) & ViT-B-16 & 94.47 & 98.90 & 96.35 & 97.37 & 91.98 & 85.68 \\
IN-200 & ResNet-18 (scr) & ViT-L-14 & 96.33 & 99.24 & 97.32 & 98.41 & 94.45 & 88.52 \\
IN-200 & ResNet-18 (scr) & ViT-H-14 & 96.28 & 99.00 & 97.36 & 97.96 & 93.97 & 87.79 \\
IN-200 & ResNet-18 (ft) & ViT-B-32 & 92.42 & 98.30 & 95.64 & 96.29 & 90.48 & 83.28 \\
IN-200 & ResNet-18 (ft) & ViT-B-16 & 94.52 & 98.85 & 96.35 & 97.35 & 91.85 & 85.63 \\
IN-200 & ResNet-18 (ft) & ViT-L-14 & 96.28 & 99.21 & 97.30 & 98.38 & 94.36 & 88.47 \\
IN-200 & ResNet-18 (ft) & ViT-H-14 & 96.33 & 98.95 & 97.36 & 97.94 & 93.86 & 87.74 \\
IN-200 & ResNet-50 (ft) & ViT-B-32 & 93.61 & 98.51 & 95.99 & 96.62 & 91.06 & 83.34 \\
IN-200 & ResNet-50 (ft) & ViT-B-16 & 95.17 & 98.92 & 96.67 & 97.57 & 92.38 & 85.80 \\
IN-200 & ResNet-50 (ft) & ViT-L-14 & 96.57 & 99.27 & 97.60 & 98.52 & 94.77 & 88.97 \\
IN-200 & ResNet-50 (ft) & ViT-H-14 & 96.41 & 99.04 & 97.79 & 98.21 & 94.64 & 88.47 \\
IN-200 & ViT-B16 (ft) & ViT-B-32 & 92.68 & 98.37 & 96.04 & 96.40 & 90.61 & 82.75 \\
IN-200 & ViT-B16 (ft) & ViT-B-16 & 94.49 & 98.87 & 96.76 & 97.46 & 92.05 & 85.35 \\
IN-200 & ViT-B16 (ft) & ViT-L-14 & 96.26 & 99.26 & 97.69 & 98.49 & 94.64 & 88.54 \\
IN-200 & ViT-B16 (ft) & ViT-H-14 & 96.40 & 99.03 & 97.83 & 98.13 & 94.37 & 87.96 \\
\bottomrule
\end{tabular}%
}

\vspace{2em}\resizebox{0.9\linewidth}{!}{%
\begin{tabular}{lllrrrrrr}
\toprule
Dataset & Classifier & CLIP & ID Acc. & iNaturalist & Textures & OpenImage-O & NINCO & SSB-Hard \\
\midrule
IN-200 (CS) & ResNet-18 (scr) & ViT-B-32 & 71.80 & 83.65 & 77.85 & 76.19 & 61.08 & 52.20 \\
IN-200 (CS) & ResNet-18 (scr) & ViT-B-16 & 78.08 & 88.34 & 80.78 & 80.96 & 64.57 & 56.41 \\
IN-200 (CS) & ResNet-18 (scr) & ViT-L-14 & 88.07 & 93.07 & 87.32 & 88.54 & 75.38 & 65.78 \\
IN-200 (CS) & ResNet-18 (scr) & ViT-H-14 & 89.64 & 90.46 & 86.80 & 85.33 & 72.10 & 62.71 \\
IN-200 (CS) & ResNet-18 (ft) & ViT-B-32 & 72.03 & 83.30 & 78.19 & 76.22 & 60.86 & 52.38 \\
IN-200 (CS) & ResNet-18 (ft) & ViT-B-16 & 78.24 & 88.00 & 81.00 & 80.95 & 64.36 & 56.56 \\
IN-200 (CS) & ResNet-18 (ft) & ViT-L-14 & 88.00 & 92.84 & 87.46 & 88.49 & 75.15 & 65.85 \\
IN-200 (CS) & ResNet-18 (ft) & ViT-H-14 & 89.64 & 90.17 & 86.98 & 85.33 & 71.83 & 62.84 \\
IN-200 (CS) & ResNet-50 (ft) & ViT-B-32 & 72.35 & 82.92 & 77.60 & 76.25 & 61.46 & 52.22 \\
IN-200 (CS) & ResNet-50 (ft) & ViT-B-16 & 78.09 & 87.54 & 80.52 & 80.81 & 64.81 & 56.29 \\
IN-200 (CS) & ResNet-50 (ft) & ViT-L-14 & 87.87 & 92.51 & 87.16 & 88.27 & 75.29 & 65.48 \\
IN-200 (CS) & ResNet-50 (ft) & ViT-H-14 & 89.47 & 89.74 & 86.70 & 85.19 & 72.23 & 62.51 \\
IN-200 (CS) & ViT-B16 (ft) & ViT-B-32 & 70.53 & 82.85 & 79.37 & 76.22 & 60.72 & 51.31 \\
IN-200 (CS) & ViT-B16 (ft) & ViT-B-16 & 76.80 & 87.63 & 82.18 & 80.97 & 64.23 & 55.58 \\
IN-200 (CS) & ViT-B16 (ft) & ViT-L-14 & 87.22 & 92.65 & 88.51 & 88.55 & 75.27 & 65.20 \\
IN-200 (CS) & ViT-B16 (ft) & ViT-H-14 & 89.15 & 89.92 & 88.02 & 85.41 & 72.01 & 62.05 \\
\bottomrule
\end{tabular}%
}
\caption{COOkeD performance on ImageNet-200 (standard and full-spectrum evaluation).}
\end{table}

\begin{table}
\centering
\vspace{2em}\resizebox{0.9\linewidth}{!}{%
\begin{tabular}{lllrrrrrr}
\toprule
Dataset & Classifier & CLIP & ID Acc. & iNaturalist & Textures & OpenImage-O & NINCO & SSB-Hard \\
\midrule
IN-1K & ResNet-18 (ft) & ViT-B-32 & 74.56 & 94.63 & 88.43 & 90.76 & 80.12 & 64.91 \\
IN-1K & ResNet-18 (ft) & ViT-B-16 & 77.26 & 95.97 & 90.16 & 92.88 & 82.10 & 68.23 \\
IN-1K & ResNet-18 (ft) & ViT-L-14 & 81.83 & 97.64 & 91.62 & 95.62 & 87.67 & 73.61 \\
IN-1K & ResNet-18 (ft) & ViT-H-14 & 82.63 & 96.84 & 92.38 & 94.94 & 88.28 & 74.21 \\
IN-1K & ResNet-50 (ft) & ViT-B-32 & 77.85 & 95.05 & 88.93 & 91.69 & 81.04 & 66.21 \\
IN-1K & ResNet-50 (ft) & ViT-B-16 & 79.76 & 96.24 & 90.59 & 93.61 & 82.95 & 69.62 \\
IN-1K & ResNet-50 (ft) & ViT-L-14 & 82.99 & 97.76 & 91.91 & 96.01 & 88.37 & 75.09 \\
IN-1K & ResNet-50 (ft) & ViT-H-14 & 83.68 & 97.01 & 92.76 & 95.45 & 89.15 & 75.96 \\
IN-1K & ViT-B16 (ft) & ViT-B-32 & 80.85 & 95.08 & 89.43 & 91.83 & 80.82 & 65.36 \\
IN-1K & ViT-B16 (ft) & ViT-B-16 & 82.21 & 96.20 & 91.00 & 93.68 & 82.85 & 68.97 \\
IN-1K & ViT-B16 (ft) & ViT-L-14 & 84.57 & 97.71 & 92.19 & 96.03 & 88.42 & 74.69 \\
IN-1K & ViT-B16 (ft) & ViT-H-14 & 84.73 & 97.00 & 93.07 & 95.44 & 89.27 & 75.38 \\
\bottomrule
\end{tabular}%
}

\vspace{2em}\resizebox{0.9\linewidth}{!}{%
\begin{tabular}{lllrrrrrr}
\toprule
Dataset & Classifier & CLIP & ID Acc. & iNaturalist & Textures & OpenImage-O & NINCO & SSB-Hard \\
\midrule
IN-1K (CS) & ResNet-18 (ft) & ViT-B-32 & 60.71 & 83.25 & 74.21 & 76.53 & 61.91 & 47.19 \\
IN-1K (CS) & ResNet-18 (ft) & ViT-B-16 & 64.67 & 85.80 & 76.64 & 79.75 & 63.64 & 49.72 \\
IN-1K (CS) & ResNet-18 (ft) & ViT-L-14 & 73.73 & 91.27 & 80.46 & 86.48 & 72.34 & 56.24 \\
IN-1K (CS) & ResNet-18 (ft) & ViT-H-14 & 76.41 & 88.36 & 81.26 & 84.25 & 72.22 & 56.58 \\
IN-1K (CS) & ResNet-50 (ft) & ViT-B-32 & 63.03 & 82.96 & 74.05 & 77.20 & 62.32 & 48.20 \\
IN-1K (CS) & ResNet-50 (ft) & ViT-B-16 & 66.17 & 85.41 & 76.41 & 80.27 & 63.96 & 50.82 \\
IN-1K (CS) & ResNet-50 (ft) & ViT-L-14 & 74.33 & 90.86 & 80.22 & 86.68 & 72.52 & 57.40 \\
IN-1K (CS) & ResNet-50 (ft) & ViT-H-14 & 76.93 & 87.90 & 81.08 & 84.54 & 72.50 & 57.94 \\
IN-1K (CS) & ViT-B16 (ft) & ViT-B-32 & 66.73 & 83.52 & 75.68 & 78.34 & 63.23 & 48.21 \\
IN-1K (CS) & ViT-B16 (ft) & ViT-B-16 & 69.30 & 85.87 & 77.93 & 81.29 & 64.91 & 50.97 \\
IN-1K (CS) & ViT-B16 (ft) & ViT-L-14 & 76.08 & 91.09 & 81.41 & 87.34 & 73.47 & 57.83 \\
IN-1K (CS) & ViT-B16 (ft) & ViT-H-14 & 78.16 & 88.30 & 82.28 & 85.30 & 73.73 & 58.38 \\
\bottomrule
\end{tabular}%
}

\caption{COOkeD performance on ImageNet-1K (standard and full-spectrum evaluation).}
\end{table}

\end{document}